\documentclass[11pt]{article}

\newif\ifpreprint
\preprinttrue
\ifpreprint
  \usepackage[preprint]{acl}
\else
  \usepackage[review]{acl}
\fi

\usepackage{fontspec}
\usepackage{xeCJK}
\setCJKsansfont{FandolHei-Regular}[Extension = .otf, BoldFont = FandolHei-Bold]
\setCJKmonofont{FandolFang-Regular}[Extension = .otf]

\usepackage{microtype}
\usepackage{amsmath,amssymb}
\usepackage{graphicx}
\usepackage{booktabs}
\usepackage{multirow}
\usepackage{xcolor}
\usepackage{enumitem}
\usepackage{fvextra}
\usepackage{xspace}
\setlist{nosep,leftmargin=*}

\newif\ifshowtodos \showtodosfalse

\newcommand{\cond}[1]{\textsc{#1}}               
\newcommand{\cat}[1]{\texttt{\detokenize{#1}}}    
\newcommand{\sheng}{\textit{sheng}\xspace}
\newcommand{\ke}{\textit{ke}\xspace}

\title{Compute the Interactions, Not the Attributes:\\What a Fact Layer Must Supply in the Chinese Sexagenary System}

\ifpreprint
  \author{Huixue Han \\ Independent Researcher \\ \texttt{ixue9@icloud.com}
    \And
    Jiaying Liu \\ Independent Researcher \\ \texttt{sammi.liu01@gmail.com}}
\else
  \author{Anonymous ACL submission}
\fi

\begin{document}
\maketitle

\begin{abstract}
Language-model systems in rule-governed domains often pair the model with a deterministic fact layer, but what that layer should compute is rarely measured. We measure it in the Chinese sexagenary (干支) system underlying BaZi (八字) chart analysis, where every fact has one table-checkable value: 682 items over twelve fact categories and 160 transit--natal collision items, each posed on a ladder of rungs that differ only in how much of the computation is done for the model. Answering directly, none of eleven open-weight models gets more than 4\% of sexagenary days or 1\% of collisions right, and supplying the attributes of every pillar does not change the collisions. A complete list of computed relations is not enough either: it leaves Qwen3-32B at 1\%, with relations listed for other pairs misattributed to the queried pair, whereas a query-filtered layer from follow-up runs, stating only the queried relations pillar by pillar, gives 92\%. On chess positions, two predictions locked in advance hold: attributes do not help, and unqueried relations are again misattributed. Three proprietary reasoning models compute 68--95\% of collisions unaided, while reasoning in smaller open models overrides supplied facts. A fact layer should compute the interactions and return the ones the question asks about.
\end{abstract}

\section{Introduction}
\label{sec:intro}

\begin{figure*}[t]
\centering
\includegraphics[width=\textwidth]{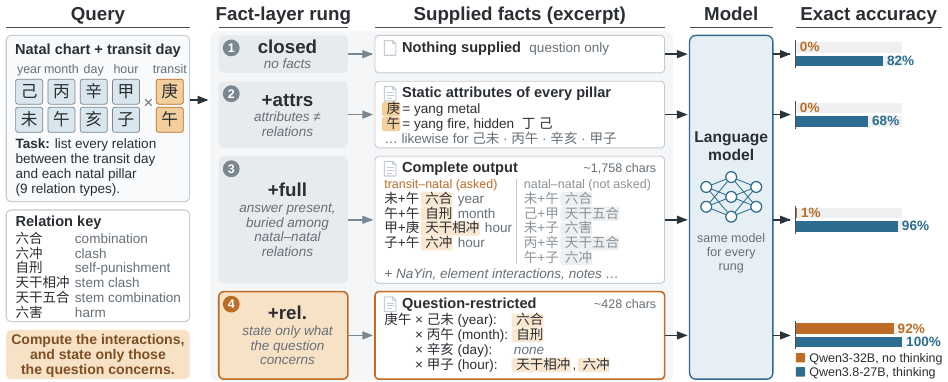}
\caption{The fact-layer ladder on one collision item (transit pillar given). Each rung supplies more of the computation to the same model; bars give exact accuracy over 80 items (orange: Qwen3-32B without thinking; blue: Qwen3.8-27B with thinking). Static attributes help neither model. The complete list (\cond{+full}) contains every relation the answer needs, mixed with relations between natal pillars that the question does not concern, and Qwen3-32B answers 1\% of items from it; stated per natal pillar by the query-filtered layer (\cond{+rel.}), the same relations give 92\%.}
\label{fig:framework}
\end{figure*}

Tool-augmented and retrieval-augmented systems supply a language model with computed facts or retrieved rules \citep{gao2023pal,schick2023toolformer,lewis2020rag}. In a rule-governed domain, the designer of such a component chooses what it returns: the attributes of each entity, every relation among the entities, or only the relation the user asked about. Returning attributes is the simplest choice and leaves their combination to the model. Prior work shows that long or noisy tool output costs accuracy and that compressing it can help \citep{kate2025longfunceval,lele2026tooltrim}; it does not compare these three outputs on the same questions. The comparison matters (Figure~\ref{fig:framework}): answering directly, Qwen3-32B gets none of one item set right with the attributes of every entity, and 1\% with a complete list of relations that contains the answer, misattributing relations listed for other pairs to the pair it was asked about.

We measure this in the Chinese sexagenary (干支, \textit{ganzhi}) system. Its sixty-term cycle of stems and branches has dated days since the late Shang period \citep{keightley2000ancestral,wilkinson2017chinese}, and its tables of elements and pairwise relations are the factual layer of BaZi (八字, Four Pillars) chart analysis. A reading reasons in steps such as ``the day master is 癸 water; the year stem 丙 is fire, which 癸 controls, so it is the wealth star; today is a 丁丑 day, and 丁 clashes with 癸''. Every clause before the interpretation is a look-up or a short derivation over fixed tables, and when one is wrong the reading stays fluent. The domain suits the design question: every fact has one value once a convention is fixed, facts range from single look-ups to combinations of several attributes, and the chart separates cleanly from interpretation. Prior BaZi benchmarks score interpretive or predictive answers whose ground truth is contested \citep{zheng2025bazi,chen2026baziqa}; we score only table-checkable facts and vary how much of the computation is done for the model.

We ask three questions. \textbf{RQ1}: in which categories of table-checkable facts do models err without help? \textbf{RQ2}: does the reference table in context remove these errors? \textbf{RQ3}: which parts of the computation must a fact layer perform for the errors to disappear? Our contributions are:
\begin{enumerate}
\item \textbf{A benchmark with a fact-layer ladder}: 682 items over twelve categories and 160 transit--natal collision items, posed under conditions that differ only in what is computed for the model, with gold answers reproduced by an independent implementation of the rules and checked exhaustively against answer-set programs (\S\ref{sec:gold}).
\item \textbf{Errors concentrate in the day count and in collisions}: stem elements are at 100\% for every Qwen model from 4B and for Gemma, while answering directly every open-weight model is at or below the majority baseline on closed-book day pillars and at 0--1\% on collisions between pillars; models from 4B upward name 34--84\% of weekdays on the same dates (\S\ref{sec:calendar}).
\item \textbf{Attributes do not substitute for relations}: the polarity, element and hidden stems of every pillar have no consistent effect on collisions. On chess positions in English, where relations do follow from attributes, the same prediction, locked and timestamped before any run, holds in all 18 decidable cells (\S\ref{sec:collision}, \S\ref{sec:chess}).
\item \textbf{A complete list is not yet a readable one}: the complete list states every relation an answer needs, yet leaves every Qwen3 model answering directly at 0--1\%, and relations it lists between other pairs are misattributed to the queried pair (Qwen3-32B: 34.9\% against 10.3\% for unlisted types). The same misattribution is significant in 11 of 13 chess runs under a locked prediction. In follow-up runs, a query-filtered layer that states only the queried relations raises every model from 1.7B upward to 32--100\% (Qwen3-32B: 90--92\%; four models: 96--100\%; \S\ref{sec:collision}).
\item \textbf{What reasoning changes}: with extended reasoning, models use the reference table, the two Qwen3.8 models count days, and three proprietary reasoning models compute 68--95\% of collisions unaided and 99--100\% with the complete list, so the choice of layer matters most for models that answer directly. In Qwen3 models up to 8B, reasoning also re-derives supplied facts against an explicit instruction not to, lowering accuracy by up to 31 points, so systems should check answers against the fact layer as well as condition on it (\S\ref{sec:static}, \S\ref{sec:confirmatory}).
\end{enumerate}

\section{Related Work}
\label{sec:related}

\paragraph{BaZi and language models.}
\citet{zheng2025bazi} predict life events of 50 public figures with a system that combines symbolic BaZi features with a language model; \citet{chen2026baziqa} collect 200 multiple-choice problems from a professional BaZi competition and supply a pre-computed chart to separate calendar conversion from inference. \citet{li2026knowing} pose 3,000 multiple-choice questions, with a model-generated answer key and no tools, and find that theory questions score above chart-based cases for all six systems. These studies score interpretive or predictive answers.

\paragraph{Temporal and compositional reasoning.}
Date arithmetic is a known weakness of language models \citep{tan2023towards,chu2024timebench,fatemi2025testoftime}, including conversion between the Gregorian and Chinese lunar calendars \citep{miao2026span}. Models answer sub-questions they cannot combine \citep{press2023compositionality,dziri2023faith}; in a transit--natal collision every sub-fact is a look-up and the difficulty lies in the pairwise combination.

\paragraph{Externalising computation.}
Delegating computation to programs \citep{gao2023pal} or tools \citep{schick2023toolformer} and retrieving evidence into the context \citep{lewis2020rag,vu2024freshllms} are the standard remedies for such errors; our \cond{table} condition corresponds to retrieving the rules and our \cond{tool} conditions to program-computed facts.

\paragraph{Tool output and distractors.}
Answer retrieval degrades as tool responses grow \citep{kate2025longfunceval}, and query-aware pruning or compression of tool output can keep accuracy while removing most tokens \citep{kovacs2026squeez,lele2026tooltrim}. \citet{lele2026tooltrim} also report that removing noise from bloated output lifts two frontier models from 73\% to 92--100\%. \citet{zhang2026same} change how a simulator's report reaches the agent with its content fixed. These studies vary how much of an output remains or how it is delivered; we vary which layer of computed facts is returned. Irrelevant sentences and documents in a problem or retrieved context lower accuracy \citep{shi2023distracted,yoran2024robust,yang2025distracted,lee2026noisybench}. Closest to our mechanism, irrelevant rules harm models in RuleArena, while meaningless padding of the same length costs little \citep{zhou2025rulearena}. In \cond{+full} the distractors are true relations printed by the fact layer itself. We add chess as a second rule system; earlier work uses it to study state tracking and policy learning \citep{toshniwal2022chess,feng2023chessgpt}.

\section{Benchmark}
\label{sec:benchmark}

\subsection{Reference table and gold answers}
\label{sec:gold}

All gold answers are generated from a single JSON table that serves as the fact layer of an open-source BaZi assistant skill.\footnote{Repository and commit withheld for review; the table is versioned as \texttt{table.json v1}.} It contains the polarity and element of the ten stems and twelve branches, the hidden stems of each branch, the \sheng (generating) and \ke (controlling) cycles of the five elements, the ten-god rules, the seasonal strength states, the sixty-pillar NaYin table, the stem combinations and clashes, and seven branch relation types (六合 combination, 半合 half combination, 六冲 clash, 相刑 punishment, 六害 harm, 六破 destruction, 自刑 self-punishment). Day pillars are counted from the anchor 1949-10-01 = 甲子, with the civil day changing at 00:00.

Where schools differ, the table fixes one convention and every affected question states it: 午未 combine to earth; a half combination requires the middle branch 子, 午, 卯 or 酉; the four 季 months count as earth-ruled; only pairwise relations are scored. A second implementation written from standard reference tables, sharing no code with the table, reproduces every gold answer of all 842 items, including the relation labels of the 160 collision items. The calendar library \texttt{lunar\_python} 1.4.8 agrees on all 392 answer fields it can derive. Answer-set programs that define the relations arithmetically agree with both implementations on every case of their finite domains, including every date from 1900 to 2100 (Appendix~\ref{app:formal}).

\subsection{Fact categories}
\label{sec:categories}

Table~\ref{tab:categories} lists the twelve categories, from single look-ups (the element of a stem) through derivations over the same tables (a ten-god label needs the element relation, its direction and the polarities of two stems) to date arithmetic (the day pillar of a Gregorian date) and a four-step chain (date $\to$ transit day pillar $\to$ ten god against the natal day master). Answers are requested as JSON under a per-item schema whose enumerations list the admissible labels, so formatting errors are not counted as hallucinations; NaYin and pillars remain free text. Each category is reported next to its majority-class baseline.

\begin{table}[t]
\centering\footnotesize
\setlength{\tabcolsep}{3pt}
\begin{tabular}{lrll}
\toprule
Abbr. & $n$ & Steps & What is asked \\
\midrule
StE & 10 & 1 & polarity, element of a stem \\
BrE & 12 & 1 & polarity, element of a branch \\
ERel & 45 & 2 & \textit{sheng}/\textit{ke} relation, acting side \\
TG & 100 & 3 & ten-god label vs.\ a day master \\
Hid & 12 & 1 & ordered hidden stems of a branch \\
NaY & 60 & 1 & NaYin of a pillar (free text) \\
StR & 45 & 1 & stem combinations and clashes \\
BrR & 78 & 1--2 & seven branch relation types \\
Sea & 60 & 2 & seasonal strength of an element \\
DP & 100 & date & day pillar of a date (free text) \\
Abs & 100 & 2 & absent elements, surface vs.\ hidden \\
TrTG & 60 & 4 & date $\to$ pillar $\to$ ten god vs.\ natal \\
\midrule
Total & 682 & & \\
\bottomrule
\end{tabular}
\caption{The twelve fact categories. Steps: table look-ups or derivation steps from the reference table to the gold answer.}
\label{tab:categories}
\end{table}

\subsection{Conditions}
\label{sec:conditions}

Every item is posed in three conditions that differ only in the context: \cond{closed}, the question only; \cond{table}, the full reference table with the ten-god and day-counting rules (about 1.7k characters) in the system prompt, from which the model must find and apply the relevant rows; and \cond{tool}, the question preceded by the fact lines that the deterministic script prints for the relevant stems, branches, pillars or date, under the skill's ``CONFIRMED BY CODE -- DO NOT RECALCULATE'' header. Prompts are in Simplified Chinese (Appendix~\ref{app:prompts}).

\subsection{Transit--natal collisions}
\label{sec:collision-set}

Practical use asks a combined question: how does a given day interact with a person's chart? We draw 80 natal charts, sampling each of the four pillars independently from the sixty terms, and pair each chart with a date (seed 20261031). The model must state the transit day pillar, list for each of the four natal pillars every pairwise relation it forms with the transit pillar (from nine stem and branch relation types), and give the ten god of the transit stem. An item is correct only if all four relation sets are exactly right. Each pair appears in a \emph{date given} variant, in which the model must first find the transit pillar, and a \emph{pillar given} variant. The condition ladder is extended by the amount of computation supplied: \cond{closed}, the question only; \cond{+pillar} (date variant only), the computed transit pillar; \cond{+attrs}, polarity, element and hidden stems of every pillar, without ten gods or relations; \cond{+full}, the complete list: the fact layer's whole output, which lists all relations in the chart, including relations among the natal pillars that the question does not concern; \cond{table}, the reference table instead of computed facts; and \cond{+rel.}, a query-filtered layer that states only the transit pillar, the ten god of its stem, and the relations between the transit day and each natal pillar (``none'' where there is none). This gives 880 requests per model and mode. \cond{+attrs} versus \cond{+full} measures the value of computing interactions. \cond{+full} versus \cond{+rel.} compares the complete list with the query-filtered layer, which differs from it in content, length and layout (\S\ref{sec:collision}). \cond{+rel.} states the answer slot by slot and assumes a perfect filter, so it measures whether a model can read a stated answer; \cond{+full} measures whether it can find that answer among true relations it was not asked about.

\subsection{A second domain: chess}
\label{sec:chess-set}

To test whether the design result holds for a different rule system and language, we built a chess set with the same structure. Each item places four pieces on a board, the counterpart of the natal pillars, and one new piece, the counterpart of the transit day. For each of the four pieces the model lists every relation the new piece forms with it, from seven types (attacks, attacked by, defends, defended by, same file, same rank, same diagonal). Every board contains at least one attack or defence between the new piece and a piece on the board, and one between two pieces already on it, the counterpart of a natal--natal relation. The 160 items (seed 20261105) are posed in English under six conditions that mirror the collision ladder: \cond{closed}; \cond{table}, with movement and relation definitions; \cond{+attrs}, with colour, piece type, square, file, rank and square colour of every piece; \cond{+full}, with all attributes and every pairwise relation on the board; \cond{+full$-$natal}; and \cond{+rel.}. Gold answers come from a standard-library implementation and agree with python-chess 1.11.2 on all 640 answer slots and on every relation line of \cond{+full}.

\section{Experimental Setup}
\label{sec:setup}

\paragraph{Models.}
We evaluate eleven open-weight instruction-tuned models: the Qwen3 dense series at 0.6B, 1.7B, 4B, 8B, 14B and 32B parameters; three later Qwen releases (Qwen3.6-35B-A3B, a mixture-of-experts model with 3B active parameters; Qwen3.8-27B; Qwen3.8-Flash-Next in FP8); Gemma-4-31B-it; and OLMo-3.1-32B-Instruct \citep{yang2025qwen3,gemma4card,olmo31card} (revisions in Appendix~\ref{app:repro}). Models that support it are run without extended reasoning (``no thinking'', answering directly) and with it (``thinking'').

\paragraph{Inference.}
All runs use vLLM 0.32.0 \citep{kwon2023vllm} with JSON-schema structured outputs. No-thinking runs use greedy decoding with a 512-token cap, and the system prompt asks for the facts only, without explanation, so they measure direct answers. Thinking runs use each model's released sampling configuration and an 8,192-token cap, because greedy decoding with thinking produced repetition loops in a pilot. A budget probe repeats the day-pillar and transit items, and for the three strongest thinking models the most truncated collision conditions, at a 24,576-token cap. Replicate samples of selected thinking cells vary by a median of 5 points (Appendix~\ref{app:repro}).

\paragraph{Exclusions.}
Two thinking runs are excluded because their outputs do not measure the task: Gemma-4-31B-it and OLMo-3.1-32B-Think (Appendix~\ref{app:deviations}).

\paragraph{Grading and statistics.}
Free-text answers are compared with gold after trimming whitespace and mapping traditional to simplified characters; four variant spellings of NaYin names are accepted. Set-valued answers are correct only if they equal the gold set; we also report micro precision and recall over relation labels. Truncated outputs and refusals count as wrong. Because all conditions and models share items, comparisons use the exact McNemar test \citep{mcnemar1947note,dietterich1998approximate} with Holm correction \citep{holm1979simple} across the categories of a contrast.

\paragraph{Analysis plan and decision rules.}
\label{sec:protocol}
Research questions, hypotheses H1--H4 and the analysis plan were written before any model was queried. Decision rules (Appendix~\ref{app:hypotheses}) and hypotheses H5--H9 on the collision set were fixed after the open-weight runs had been inspected and before any run of the three proprietary models that form the confirmatory set (\S\ref{sec:confirmatory}); the thresholds are those of the original plan. The open-weight runs are thus the discovery set, in which the query-filtered condition and the weekday control were designed (Appendix~\ref{app:deviations}), and the proprietary runs test the resulting predictions. Predictions C1--C4 for the chess set were locked before any chess request, and the protocol and the hashes of the items, requests and grading code were timestamped with OpenTimestamps.

\section{Results}
\label{sec:results}

\begin{figure}[t]
\centering
\includegraphics[width=\columnwidth]{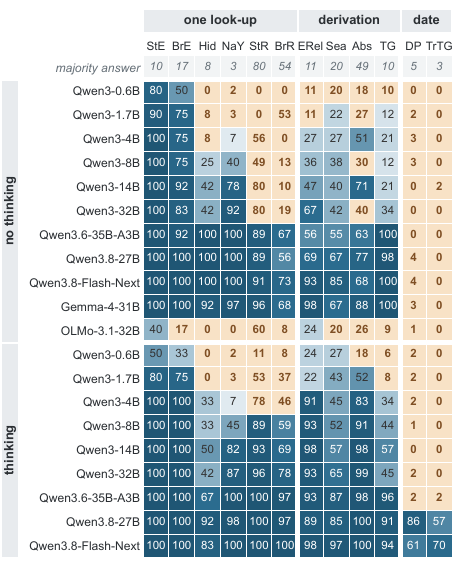}
\caption{Closed-book accuracy (\%) by category (abbreviations as in Table~\ref{tab:categories}), grouped by the steps an answer needs. Top row: majority-class answer; apricot and bold: at or below it. Look-ups rise with model generation and size; the day pillar and the transit ten god stay at or below the majority answer for every model except the two Qwen3.8 models with thinking.}
\label{fig:heatmap}
\end{figure}

Table~\ref{tab:main} summarises the open-weight runs, with the confirmatory runs (\S\ref{sec:confirmatory}) in its last block, and Figure~\ref{fig:heatmap} breaks closed-book accuracy down by category; per-category tables for every condition, the locked rules for H1--H4 applied to these runs (Table~\ref{tab:hyp-open}) and further analyses are in Appendices~\ref{app:full} and~\ref{app:extra}.

\begin{table*}[t]
\centering\small
\setlength{\tabcolsep}{3.4pt}
\begin{tabular}{lrrrrrrr@{\hspace{9pt}}rrrrrrr}
\toprule
 & \multicolumn{7}{c}{No thinking} & \multicolumn{7}{c}{Thinking} \\
\cmidrule(lr){2-8}\cmidrule(lr){9-15}
 & \multicolumn{3}{c}{Macro} & \multicolumn{2}{c}{DP} & \multicolumn{2}{c}{TrTG} & \multicolumn{3}{c}{Macro} & \multicolumn{2}{c}{DP} & \multicolumn{2}{c}{TrTG} \\
\cmidrule(lr){2-4}\cmidrule(lr){5-6}\cmidrule(lr){7-8}\cmidrule(lr){9-11}\cmidrule(lr){12-13}\cmidrule(lr){14-15}
Model & closed & table & tool & closed & table & closed & tool & closed & table & tool & closed & table & closed & tool \\
\midrule
Majority class & 23 & 23 & 23 & 5 & 5 & 3 & 3 & 23 & 23 & 23 & 5 & 5 & 3 & 3 \\
\midrule
Qwen3-0.6B & 16 & 23 & 55 & 0 & 2 & 0 & 3 & 15 & 26 & 56 & 2 & 2 & 0 & 5 \\
Qwen3-1.7B & 25 & 36 & 70 & 2 & 2 & 0 & 25 & 31 & 46 & 68 & 2 & 2 & 0 & 18 \\
Qwen3-4B & 31 & 43 & 93 & 3 & 0 & 0 & 98 & 52 & 76 & 94 & 2 & 9 & 0 & 67 \\
Qwen3-8B & 35 & 45 & 92 & 3 & 2 & 0 & 95 & 59 & 80 & 93 & 1 & 17 & 0 & 73 \\
Qwen3-14B & 49 & 54 & 100 & 0 & 0 & 2 & 97 & 67 & 81 & 97 & 0 & 14 & 0 & 85 \\
Qwen3-32B & 50 & 54 & 99 & 0 & 1 & 0 & 97 & 67 & 82 & 97 & 2 & 29 & 0 & 93 \\
\midrule
Qwen3.6-35B-A3B & 68 & 67 & 100 & 0 & 0 & 0 & 100 & 78 & 89 & 94 & 2 & 45 & 2 & 97 \\
Qwen3.8-27B & 72 & 72 & 100 & 4 & 2 & 0 & 100 & 91 & 96 & 99 & 86 & 89 & 57 & 100 \\
Qwen3.8-Flash-Next & 76 & 76 & 100 & 4 & 2 & 0 & 100 & 92 & 99 & 99 & 61 & 95 & 70 & 100 \\
\midrule
Gemma-4-31B-it & 76 & 81 & 100 & 3 & 3 & 0 & 100 & -- & -- & -- & -- & -- & -- & -- \\
OLMo-3.1-32B-Inst. & 17 & 39 & 83 & 1 & 1 & 0 & 5 & -- & -- & -- & -- & -- & -- & -- \\
\midrule
Claude Haiku 5.5 & -- & -- & -- & -- & -- & -- & -- & 90 & 97 & 99 & 71 & 97 & 62 & 95 \\
Claude Sonnet 5.5 & -- & -- & -- & -- & -- & -- & -- & 98 & 100 & 100 & 96 & 99 & 92 & 100 \\
Claude Opus 5.5 & -- & -- & -- & -- & -- & -- & -- & 98 & 100 & 100 & 98 & 98 & 92 & 98 \\
\bottomrule
\end{tabular}
\caption{Accuracy (\%) on the 682-item benchmark. Macro: unweighted mean over the twelve categories; DP: day pillar; TrTG: transit ten god. No-thinking runs use greedy decoding; thinking runs use each model's released sampling configuration with an 8k-token cap; ``--'': not run or excluded (\S\ref{sec:setup}). Claude models (confirmatory set, \S\ref{sec:confirmatory}) reason at effort \texttt{low}.}
\label{tab:main}
\end{table*}

\subsection{Look-ups, the reference table and computed facts (RQ1, RQ2)}
\label{sec:static}

Stem elements are at 100\% for every Qwen3 model from 4B, all later Qwen releases and Gemma, and memorised long tables rise steeply with size (NaYin without thinking: 2\% at 0.6B, 92\% at 32B). Later releases gain more than larger ones: macro accuracy without thinking is unchanged from Qwen3-14B to Qwen3-32B (49\% and 50\%; McNemar $p=0.73$), whereas Qwen3.6-35B-A3B, with 3B active parameters, reaches 68\% ($p<10^{-29}$ against Qwen3-32B) and Qwen3.8-27B 72\%.

\paragraph{The reference table helps a model that reasons over it (RQ2).}
Without thinking, the table changes 14 of 132 model--category cells significantly, 4 of them downward (Qwen3-32B on NaYin 92$\to$60\%); with thinking, it improves 26 of 108 cells and lowers none.

\paragraph{Computed facts are copied, except under reasoning.}
Computed facts improve 83 of 132 no-thinking cells and lower none (macro accuracy 92--100\% from 4B upward, except OLMo at 83\%). Thinking lowers six \cond{tool} cells by up to 31 points (Qwen3-4B transit ten god 98$\to$67\%): the trace re-derives a supplied fact and overrides it, despite the instruction not to recalculate.

\subsection{The day pillar needs an anchor and a long count (RQ1)}
\label{sec:calendar}

Without thinking, every model answers 0--4\% of closed-book day pillars correctly, at or below the 5\% majority baseline, and 0--2\% of transit ten gods. With thinking, the two Qwen3.8 models compute the day pillar (Qwen3.8-27B 86\%, Qwen3.8-Flash-Next 61\%) and reach 92\% and 87\% at a 24k-token cap (Table~\ref{tab:budget}). Their thinking traces, graded step by step, show where the others fail (Table~\ref{tab:cot}): the Qwen3 models from 4B to 32B state a correct date--pillar pair in at most 1\% of closed-book traces; Qwen3.6 states mostly wrong ones (9\% correct), and at a 24k cap 39\% of its wrong answers follow one of them under a correct count (6\% by chance); the two Qwen3.8 models state a correct pair in 99--100\% of traces and the correct count in 94--97\%.
\paragraph{Weekdays on the same dates do not show the failure.}
On the weekday of the same 100 dates (Table~\ref{tab:weekday}), models from 4B upward answer 34--84\% correctly without thinking, against 0--4\% of day pillars (majority answer 21\%), in 7--12 output tokens, the length of the JSON answer itself; their errors are mostly one day off (62--96\%, against 33\% for a uniformly drawn wrong answer). Such short answers and near misses suggest recall of a date--weekday association learned from text rather than a count. With thinking, the same Qwen3 models state correct date--weekday pairs (63--77\% of stated pairs) and name a weekday formula in 93--99\% of traces (Table~\ref{tab:cot}).

\subsection{Collisions: attributes do not help, computed relations do (RQ3)}
\label{sec:collision}

\begin{figure*}[t]
\centering
\includegraphics[width=\textwidth]{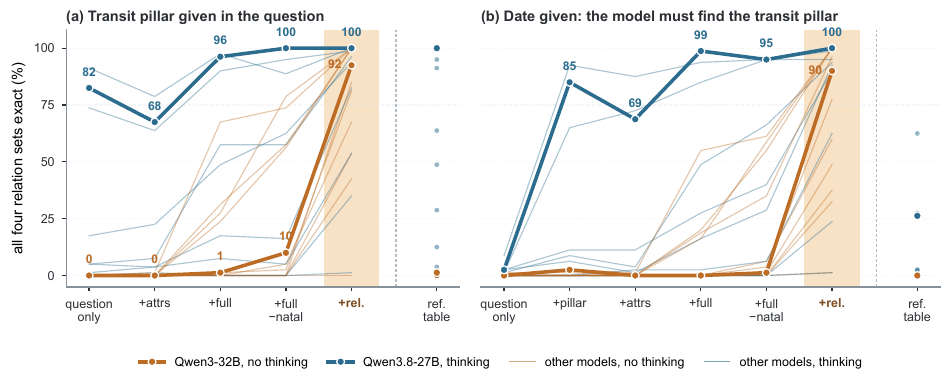}
\caption{Transit--natal collisions: share of items (\%, $n=80$ per point) whose four relation sets are all exactly right, as more of the computation is supplied (8k cap). Thin lines: every open-weight model and mode; bold: Qwen3-32B without and Qwen3.8-27B with thinking. Shaded: \cond{+rel.}; right of the grey line: the reference table instead of facts. Static attributes have no consistent effect; removing only the natal--natal relations (\cond{+full$-$natal}) recovers at most a third of what \cond{+rel.} gains for Qwen3-8B and Qwen3-32B. Full results in Tables~\ref{tab:collision} and~\ref{tab:h8}.}
\label{fig:collision}
\end{figure*}

\paragraph{Answering directly, models do not compute collisions.}
In \cond{closed} without thinking, no model gets more than 1\% of collision items fully right (0\% in 21 of 22 model--variant cells), even when the transit pillar is stated in the question; the median model has 8\% precision and 15\% recall over relation labels (Figure~\ref{fig:collision}; Table~\ref{tab:collision}).

\paragraph{Static attributes do not substitute for relations.}
The polarity, element and hidden stems of every pillar change exact accuracy by $-16$ to $+7$ points across the 40 model--mode--variant series at the 8k cap. The largest losses coincide with truncation and disappear at a 24k cap (Table~\ref{tab:collision-budget}). Combinations and clashes are identity tables over stems and branches, which these attributes do not encode; chess repeats the test where relations do follow from attributes (\S\ref{sec:chess}).

\paragraph{The complete list helps models that can select from it.}
The complete list raises the thinking Qwen3.8 models to 94--99\% and thinking Qwen3.6 to 85--90\%, but half of the 40 series stay below 10\%, including every Qwen3 model without thinking (0--1\%). The list also states relations among the natal pillars, and the larger no-thinking models misattribute these to the transit pillar. A relation type that the list states between a natal pillar and another natal pillar, but that the pillar does not form with the transit pillar, is wrongly reported for it in 34.9\% of cases by Qwen3-32B, against 10.3\% for unlisted types; for Qwen3.8-27B the rates are 16.8\% and 1.6\%.

\paragraph{A query-filtered layer is read correctly.}
In follow-up runs (\cond{+rel.} was designed after the first analysis, so they are exploratory), Qwen3-32B without thinking rises from 1\% to 92\% (pillar variant) and from 0\% to 90\% (date variant), Qwen3-8B from 1\% to 81\% and from 0\% to 78\%, and Qwen3.6, Qwen3.8-27B, Qwen3.8-Flash-Next and Gemma-4 to 96--100\%; only Qwen3-0.6B stays at 0--1\%. With thinking, Qwen3-14B and larger reach 90--100\%. The complete list states the same relations and leaves these models at 0--68\%, so the gain measures how the facts are presented, not whether they are present.

\paragraph{Where the gain comes from.}
Removing only the relation lines between two natal pillars (\cond{+full$-$natal}, H8) lowers false positives (Qwen3-32B: 4.69$\to$2.54 per item) and recovers a share $R \leq 0.31$ of the \cond{+full}$\to$\cond{+rel.} gain in all eight Qwen3-8B and Qwen3-32B cells, against 0.51--0.71 for Qwen3.8-27B without thinking (Table~\ref{tab:h8}); for the weaker readers most of the gain comes with the shorter, per-pillar layout of \cond{+rel.} (Appendix~\ref{app:extra}).

\subsection{A second domain: chess}
\label{sec:chess}

\begin{figure}[t]
\centering
\includegraphics[width=\columnwidth]{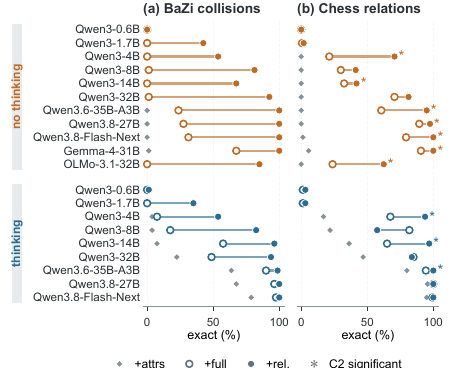}
\caption{BaZi collisions (a, transit pillar given) and chess (b) on the same ladder. Without thinking, the complete list (hollow) lifts 9 of 11 models to at least 21\% in chess, while 7 of 11 stay below 2\% in BaZi. *: \cond{+rel.} beats \cond{+full} after Holm correction (C2, chess only).}
\label{fig:chess}
\end{figure}

Figure~\ref{fig:chess} sets the chess set beside the collision set for eleven open-weight models (20 model--mode runs); Table~\ref{tab:chess} in Appendix~\ref{app:full} gives every condition. Answering directly, no model computes the relations unaided: closed-book exact accuracy is 0--1\%, and the attributes of every piece raise it to at most 6\%. C1, that attributes do not substitute for relations, holds in all 18 decidable cells, 11 of which are at 0\% under both conditions. Under \cond{+full}, a wrong slot whose relation type the list states between that piece and another piece on the board receives a false positive 1.3 to 3.0 times as often as an unlisted slot, significantly so in 11 of the 13 runs with at least 30 false positives (C4 supported; Figure~\ref{fig:chess-detail}). Two findings thus hold in both domains: attributes do not substitute for computed relations, and relations the question does not concern are misattributed to the queried pair. Unlike in the collision set, the complete list is usable when answering directly: it lifts every model from 4B upward to at least 21\%, and Gemma-4 to 91\%. Restricting it to the question adds significantly in 10 of the 18 decidable cells, for example from 21\% to 71\% for Qwen3-4B without thinking (C2 partially supported), and removing only the relations among pieces on the board accounts for less than half of that gain in 5 of 8 cells (C3 mixed; Appendix~\ref{app:deviations}). The three proprietary models score 73--97\% closed-book and 98--100\% with the complete list, so only C1 is decidable for them, for Haiku, where it holds.

\subsection{Confirmatory evaluation}
\label{sec:confirmatory}

Claude Haiku 5.5, Sonnet 5.5 and Opus 5.5 were queried at effort \texttt{low} with the prompts and answer schemas of the open-weight runs (Appendix~\ref{app:confirmatory}); median outputs of 308--1,010 tokens on closed-book day pillars show that their answers include reasoning.

\paragraph{Fact categories (H1--H4).}
Closed-book macro accuracy is 90.2\% (Haiku), 97.6\% (Sonnet) and 97.7\% (Opus); per item, error falls in that order as H4 predicts (13.0\%, 3.1\%, 1.3\%; $p_{\text{Holm}}<10^{-12}$ and $0.029$). H3 holds for all three models: \cond{tool} accuracy is at least 95\% in every category, and each of the 11 residual errors is a misquote of a fact line that states the gold answer. Sonnet and Opus compute 96--98\% of closed-book day pillars, so H4's clause that day-pillar error stays at or above 20\% fails for them and holds for Haiku (29\%). Errors concentrate in the combined categories only for Haiku (three of five; H1), and H2 is undecidable for Haiku and Sonnet and not met for Opus.

\paragraph{Collisions (H5--H9).}
Reasoning at effort \texttt{low}, the three models compute most collisions themselves: Haiku answers 86\% (pillar variant) and 68\% (date variant) of items exactly, Sonnet 95\% and 92\%, and Opus 95\% and 88\%. The complete list brings every cell to 99--100\%, and attributes move none of the three cells below ceiling by more than 5 points. The ceiling leaves the locked tests little room: \cond{+full} exceeds \cond{+attrs} by 6--10 points in the three decidable cells, short of significance under the locked correction over six cells (smallest $p_{\text{Holm}}=0.064$; H6 not supported); H5 is not decided, with three of six cells at ceiling; H7 and H8 have no decidable cell; and at most two false positives per model leave H9 unsupported. The choice between a complete and a query-filtered layer is one for models that answer directly. Appendix~\ref{app:confirmatory} reports every test and Appendix~\ref{app:deviations} the deviations from the protocol.

\section{Discussion}
\label{sec:discussion}

\paragraph{What a fact layer should compute.}
Answering directly, no model computes a collision, and supplying attributes does not help: in BaZi, attributes lie on no derivation path of a relation, and accuracy falls in tiers set by the kind of computation a condition leaves, not by its number of steps (Appendix~\ref{app:formal}); in chess, where attributes do lie on the path, C1 still holds in all 18 decidable cells. Computed relations help the models that can select from them. For models that answer directly, the query-filtered layer also removes the selection: in the follow-up collision runs it raises every model but the smallest above its accuracy with the complete list, and in chess it adds significantly in 10 of 18 cells.

\paragraph{Reasoning and scale.}
No model from 0.6B to 35B computes collisions when answering directly; the query-filtered layer brings five of the eleven to 90--100\%. Three proprietary reasoning models compute most collisions unaided, but reasoning in Qwen3 models up to 8B also re-derives supplied facts against an explicit instruction not to, so systems that pair a reasoning model with computed facts should check the answer against those facts as well as condition on them.

\paragraph{Beyond the sexagenary system.}
In both rule systems, attributes did not substitute for computed relations, and relations the question did not concern were misattributed to the queried pair. The ladder (attributes, all relations, the relations a question concerns) gives the designer of a rule-based fact component the same three-rung test: compute the interactions, and return the ones the question asks about.

\section*{Limitations}

Answers are constrained by a JSON schema with enumerated labels, which separates factual errors from formatting errors, but format constraints can themselves change performance \citep{tam2024speak}, and scoring free-form readings would additionally require extracting claims from text. The no-thinking setting combines the instruction not to explain, immediate JSON output, a 512-token cap and greedy decoding, so its results do not separate these factors and do not show what the models could do with free-text reasoning. The thinking contrast changes the decoder (vendor sampling versus greedy) as well as the reasoning mode; we bound the decoder effect without thinking. Most cells are sampled once and each condition uses one prompt template; replicates of selected thinking cells vary by a median of 5 points, a variation the McNemar tests do not include, and they generalise over items, not over runs or templates. Valid thinking runs come from one model family, because two other families did not produce usable reasoning under constrained decoding. The collision set has 80 items per cell, each in two variants, and scores pairwise relations, not three-branch combinations or punishments. Gold answers encode one set of school conventions, every affected question states its convention, and no practitioner validated the items. Dates and charts are synthetic: the four pillars are independent draws, so the month stem follows from the year stem in only 12 of the 80 charts, the hour stem from the day stem in 20, and both in 4. Pairwise relations are well defined on such charts, but their distribution may differ from that of real charts. The query-filtered layer and the weekday control were designed after the first open-weight analysis, and the query-filtered layer is tested on the collision query type with a perfect filter; a real filter would add its own errors. The weekday control also sets a cycle of 7 against one of 60. The \cond{+full$-$natal} ablation separates the natal--natal relations from the other differences between \cond{+full} and \cond{+rel.} (length, layout, explicit ``none'' lines), but not those differences from one another, and the two domains differ in language and list format as well as in rule system. The confirmatory runs cover three models from one provider at effort \texttt{low} and reach the ceiling, so they do not test the filtering predictions.

\section*{Ethics Statement}

This work measures whether language models reproduce the internal rules of the Chinese sexagenary system correctly. The sexagenary day count is a historical dating system \citep{keightley2000ancestral,wilkinson2017chinese}, and its element and relation tables are the conventions of BaZi practice; we treat both as rule systems whose internal consistency can be checked. We make no claim about the validity of BaZi as a means of prediction or character assessment. The benchmark scores agreement with a stated convention, not the truth of any reading, and we do not evaluate interpretations. All charts and dates are synthetic, no personal data are used, and no human subjects were involved.

Readings produced by assistants built on such a fact layer may influence decisions about health, finances or relationships, and more accurate facts can make these readings appear more authoritative. We recommend that deployed systems state that readings are interpretive and not predictive, and that they are not a basis for such decisions. We release the benchmark to measure and reduce factual errors of language models on a rule-defined system, not to support predictive use. The benchmark, generator, fact layer and raw outputs will be released under CC BY 4.0 (data) and the MIT licence (code).

\bibliography{references}

\appendix
\section{Prompts}
\label{app:prompts}

All prompts are in Simplified Chinese, except those of the chess set, which are in English. Each request carries a JSON schema; enumerated labels are listed in the schema. The examples below are reproduced from the request files, each followed by an English translation for the reader; the models saw only the original text. \cond{+full$-$natal} is \cond{+full} without the relation lines between two natal pillars.

\paragraph{System prompt (all conditions)}
\begin{Verbatim}[breaklines=true,breakanywhere=true,fontsize=\footnotesize,frame=single]
你在参加一个八字干支基础知识测验。只回答事实本身，严格按给定 JSON 结构作答，不要解释。
\end{Verbatim}
\noindent\textit{English translation (not shown to the models):}
\begin{Verbatim}[breaklines=true,breakanywhere=true,fontsize=\footnotesize,frame=single]
You are taking a quiz on basic BaZi stem-branch knowledge. Answer with the facts only, strictly in the given JSON structure, and do not explain.
\end{Verbatim}
\paragraph{\cond{closed}, transit ten god (user turn)}
\begin{Verbatim}[breaklines=true,breakanywhere=true,fontsize=\footnotesize,frame=single]
八字四柱为 辛未 丙辰 丁酉 乙酉（年 月 日 时）。公历 1941-10-04 这一天的流日干支（当天的日柱，不是本命日柱）是什么？用两个汉字作答。流日天干对本命日主是什么十神？
\end{Verbatim}
\noindent\textit{English translation (not shown to the models):}
\begin{Verbatim}[breaklines=true,breakanywhere=true,fontsize=\footnotesize,frame=single]
The four pillars of the BaZi chart are 辛未 丙辰 丁酉 乙酉 (year month day hour). What is the transit day pillar on the Gregorian date 1941-10-04 (the day pillar of that day, not the natal day pillar)? Answer with two Chinese characters. Which ten god is the transit-day stem relative to the natal day master?
\end{Verbatim}
\paragraph{\cond{table}: beginning of the reference table appended to the system prompt}
\begin{Verbatim}[breaklines=true,breakanywhere=true,fontsize=\footnotesize,frame=single]
下面是可供查阅的参考表：

# 干支参考表

## 天干
| 天干 | 阴阳 | 五行 |
|---|---|---|
| 甲 | 阳 | 木 |
| 乙 | 阴 | 木 |
| 丙 | 阳 | 火 |
| 丁 | 阴 | 火 |
| 戊 | 阳 | 土 |
| 己 | 阴 | 土 |
| 庚 | 阳 | 金 |
| 辛 | 阴 | 金 |
| 壬 | 阳 | 水 |
| 癸 | 阴 | 水 |

[...]
\end{Verbatim}
\noindent\textit{English translation (not shown to the models):}
\begin{Verbatim}[breaklines=true,breakanywhere=true,fontsize=\footnotesize,frame=single]
The following reference table is available for look-up:

# Stem-branch reference table

## Heavenly stems
| stem | yin/yang | element |
|---|---|---|
| 甲 | yang | wood |
| 乙 | yin | wood |
| 丙 | yang | fire |
| 丁 | yin | fire |
| 戊 | yang | earth |
| 己 | yin | earth |
| 庚 | yang | metal |
| 辛 | yin | metal |
| 壬 | yang | water |
| 癸 | yin | water |

[...]
\end{Verbatim}
\paragraph{\cond{tool}, transit ten god (user turn, fact lines truncated)}
\begin{Verbatim}[breaklines=true,breakanywhere=true,fontsize=\footnotesize,frame=single]
以下是确定性脚本计算出的事实（CONFIRMED BY CODE - DO NOT RECALCULATE, USE AS TRUTH）：

# 干支事实（CONFIRMED BY CODE - DO NOT RECALCULATE, USE AS TRUTH）
table: scripts/ganzhi/table.json v1
date: 1941-10-04
日主: 丁（阴火，俗称小火，意象灯烛星火）

## 各柱
- 年柱 辛未｜纳音 路旁土｜干 辛=阴金（小金），偏财｜支 未=阴土（羊），藏 己土(食神)、丁火(比肩)、乙木(偏印)
- 月柱 丙辰｜纳音 沙中土｜干 丙=阳火（大火），劫财｜支 辰=阳土（龙），藏 戊土(伤官)、乙木(偏印)、癸水(七杀)
- 日柱 丁酉｜纳音 山下火｜干 丁=阴火（小火），日主｜支 酉=阴金（鸡），藏 辛金(偏财)
- 时柱 乙酉｜纳音 泉中水｜干 乙=阴木（小木），偏印｜支 酉=阴金（鸡），藏 辛金(偏财)
- 流日 乙酉｜纳音 泉中水｜干 乙=阴木（小木），偏印｜支 酉=阴金（鸡），藏 辛金(偏财)

[...]
\end{Verbatim}
\noindent\textit{English translation (not shown to the models):}
\begin{Verbatim}[breaklines=true,breakanywhere=true,fontsize=\footnotesize,frame=single]
The following facts were computed by a deterministic script (CONFIRMED BY CODE - DO NOT RECALCULATE, USE AS TRUTH):

# Stem-branch facts (CONFIRMED BY CODE - DO NOT RECALCULATE, USE AS TRUTH)
table: scripts/ganzhi/table.json v1
date: 1941-10-04
Day master: 丁 (yin fire, commonly called small fire; image: lamp, candle, spark)

## Pillars
- Year pillar 辛未 | NaYin: roadside earth | stem 辛 = yin metal (small metal), Indirect Wealth | branch 未 = yin earth (goat), hidden: 己 earth (Eating God), 丁 fire (Friend), 乙 wood (Indirect Resource)
- Month pillar 丙辰 | NaYin: sand earth | stem 丙 = yang fire (big fire), Rob Wealth | branch 辰 = yang earth (dragon), hidden: 戊 earth (Hurting Officer), 乙 wood (Indirect Resource), 癸 water (Seven Killings)
- Day pillar 丁酉 | NaYin: fire at the foot of the mountain | stem 丁 = yin fire (small fire), day master | branch 酉 = yin metal (rooster), hidden: 辛 metal (Indirect Wealth)
- Hour pillar 乙酉 | NaYin: spring water | stem 乙 = yin wood (small wood), Indirect Resource | branch 酉 = yin metal (rooster), hidden: 辛 metal (Indirect Wealth)
- Transit day 乙酉 | NaYin: spring water | stem 乙 = yin wood (small wood), Indirect Resource | branch 酉 = yin metal (rooster), hidden: 辛 metal (Indirect Wealth)

[...]
\end{Verbatim}
\paragraph{Collision item, date variant, \cond{+attrs} (user turn)}
\begin{Verbatim}[breaklines=true,breakanywhere=true,fontsize=\footnotesize,frame=single]
以下是确定性脚本计算出的事实（CONFIRMED BY CODE - DO NOT RECALCULATE, USE AS TRUTH）：

- 年柱 己未｜干 己=阴土｜支 未=阴土，藏干 己土、丁火、乙木
- 月柱 丙午｜干 丙=阳火｜支 午=阳火，藏干 丁火、己土
- 日柱 辛亥｜干 辛=阴金｜支 亥=阴水，藏干 壬水、甲木
- 时柱 甲子｜干 甲=阳木｜支 子=阳水，藏干 癸水
- 流日 庚午｜干 庚=阳金｜支 午=阳火，藏干 丁火、己土

问题：八字四柱为 己未 丙午 辛亥 甲子（年 月 日 时）。公历 2009-09-22 这一天的流日干支（当天的日柱，不是本命日柱）是什么？用两个汉字作答。逐柱列出流日与该柱之间成立的全部两两关系，从 天干五合、天干相冲、六合、半合、六冲、相刑、六害、六破、自刑 中选择（只做两两比较，不计三合、三会、三刑全组；半合只算含子午卯酉中神的两支），都不成立则返回空列表。流日天干对本命日主是什么十神？
\end{Verbatim}
\noindent\textit{English translation (not shown to the models):}
\begin{Verbatim}[breaklines=true,breakanywhere=true,fontsize=\footnotesize,frame=single]
The following facts were computed by a deterministic script (CONFIRMED BY CODE - DO NOT RECALCULATE, USE AS TRUTH):

- Year pillar 己未 | stem 己 = yin earth | branch 未 = yin earth, hidden stems 己 earth, 丁 fire, 乙 wood
- Month pillar 丙午 | stem 丙 = yang fire | branch 午 = yang fire, hidden stems 丁 fire, 己 earth
- Day pillar 辛亥 | stem 辛 = yin metal | branch 亥 = yin water, hidden stems 壬 water, 甲 wood
- Hour pillar 甲子 | stem 甲 = yang wood | branch 子 = yang water, hidden stems 癸 water
- Transit day 庚午 | stem 庚 = yang metal | branch 午 = yang fire, hidden stems 丁 fire, 己 earth

Question: The four pillars of the BaZi chart are 己未 丙午 辛亥 甲子 (year month day hour). What is the transit day pillar on the Gregorian date 2009-09-22 (the day pillar of that day, not the natal day pillar)? Answer with two Chinese characters. For each pillar, list every pairwise relation that holds between the transit day and that pillar, choosing from stem combination (天干五合), stem clash (天干相冲), combination (六合), half combination (半合), clash (六冲), punishment (相刑), harm (六害), destruction (六破) and self-punishment (自刑) (pairwise comparisons only: three-branch combinations, directional combinations and complete three-way punishments are not counted; a half combination counts only for two branches that include a middle branch 子, 午, 卯 or 酉). Return an empty list if none holds. Which ten god is the transit-day stem relative to the natal day master?
\end{Verbatim}
\paragraph{Collision item, \cond{+full}: relation block of the fact-layer output}
\begin{Verbatim}[breaklines=true,breakanywhere=true,fontsize=\footnotesize,frame=single]
- [支] 六合 → 土：年柱未 + 月柱午
- [干] 天干五合 → 土：年柱己 + 时柱甲
- [支] 六害：年柱未 + 时柱子
- [支] 六合 → 土：年柱未 + 流日午
- [干] 天干五合 → 水：月柱丙 + 日柱辛
- [支] 六冲：月柱午 + 时柱子
- [支] 自刑：月柱午 + 流日午
- [干] 天干相冲：时柱甲 + 流日庚
- [支] 六冲：时柱子 + 流日午
\end{Verbatim}
\noindent\textit{English translation (not shown to the models):}
\begin{Verbatim}[breaklines=true,breakanywhere=true,fontsize=\footnotesize,frame=single]
- [branch] combination (六合) → earth: year pillar 未 + month pillar 午
- [stem] stem combination (天干五合) → earth: year pillar 己 + hour pillar 甲
- [branch] harm (六害): year pillar 未 + hour pillar 子
- [branch] combination (六合) → earth: year pillar 未 + transit day 午
- [stem] stem combination (天干五合) → water: month pillar 丙 + day pillar 辛
- [branch] clash (六冲): month pillar 午 + hour pillar 子
- [branch] self-punishment (自刑): month pillar 午 + transit day 午
- [stem] stem clash (天干相冲): hour pillar 甲 + transit day 庚
- [branch] clash (六冲): hour pillar 子 + transit day 午
\end{Verbatim}
\paragraph{Collision item, date variant, \cond{+rel.}: fact block}
\begin{Verbatim}[breaklines=true,breakanywhere=true,fontsize=\footnotesize,frame=single]
以下是确定性脚本计算出的事实（CONFIRMED BY CODE - DO NOT RECALCULATE, USE AS TRUTH）：

- 流日（公历 2009-09-22）干支：庚午
- 流日天干 庚=阳金，对本命日主 辛 为 劫财
- 流日与本命各柱之间成立的两两关系（只列流日与该柱，不含本命各柱之间的关系）：
  - 流日 庚午 × 年柱 己未：六合
  - 流日 庚午 × 月柱 丙午：自刑
  - 流日 庚午 × 日柱 辛亥：无
  - 流日 庚午 × 时柱 甲子：天干相冲、六冲
\end{Verbatim}
\noindent\textit{English translation (not shown to the models):}
\begin{Verbatim}[breaklines=true,breakanywhere=true,fontsize=\footnotesize,frame=single]
The following facts were computed by a deterministic script (CONFIRMED BY CODE - DO NOT RECALCULATE, USE AS TRUTH):

- Transit day (Gregorian 2009-09-22) pillar: 庚午
- Transit stem 庚 = yang metal; relative to the natal day master 辛 it is rob wealth (劫财)
- Pairwise relations between the transit day and each natal pillar (transit day and that pillar only, not relations among natal pillars):
  - Transit 庚午 x year pillar 己未: combination (六合)
  - Transit 庚午 x month pillar 丙午: self-punishment (自刑)
  - Transit 庚午 x day pillar 辛亥: none
  - Transit 庚午 x hour pillar 甲子: stem clash (天干相冲), clash (六冲)
\end{Verbatim}
\paragraph{Chess item, system prompt and \cond{closed} question (original English)}
\begin{Verbatim}[breaklines=true,breakanywhere=true,fontsize=\footnotesize,frame=single]
You are taking a chess facts quiz. State only the facts, answer strictly in the given JSON structure, and do not explain.

Four pieces stand on a chessboard, and no other pieces are on it. Piece 1: white pawn on c2; Piece 2: black knight on c7; Piece 3: black queen on h6; Piece 4: white queen on a5. A black rook on a8 is now placed on the board (the arriving piece). For each of the four pieces, list every relation that holds between the arriving piece and that piece, chosen from attacks, attacked_by, defends, defended_by, same_file, same_rank, same_diagonal, and return an empty list if none holds. Relations are read from the arriving piece's point of view: attacks means the arriving piece could capture that piece (opposite colours); defends means it protects that piece (same colour); attacked_by and defended_by mean that piece attacks or protects the arriving piece. Pawns cover only the two squares diagonally forward (white moves towards rank 8); rooks, bishops and queens are blocked by any piece in between; knights jump. Ignore check, pins and whose turn it is. same_file, same_rank and same_diagonal hold when the two squares share a file, a rank or a diagonal, whatever stands between them.
\end{Verbatim}
\paragraph{Chess item, \cond{+full}: fact block}
\begin{Verbatim}[breaklines=true,breakanywhere=true,fontsize=\footnotesize,frame=single]
The following facts were computed by a deterministic script (CONFIRMED BY CODE - DO NOT RECALCULATE, USE AS TRUTH):

Pieces on the board:
- Piece 1: white pawn on c2 | file c (3), rank 2 | light square
- Piece 2: black knight on c7 | file c (3), rank 7 | dark square
- Piece 3: black queen on h6 | file h (8), rank 6 | dark square
- Piece 4: white queen on a5 | file a (1), rank 5 | dark square
- Arriving piece: black rook on a8 | file a (1), rank 8 | light square
Pairwise relations (every pair of pieces on the board, read from the first piece's point of view; pairs without a relation are omitted):
- Piece 1 (c2) - Piece 2 (c7): same_file
- Piece 2 (c7) - Piece 4 (a5): attacked_by, same_diagonal
- Arriving piece (a8) - Piece 2 (c7): defended_by
- Arriving piece (a8) - Piece 4 (a5): attacks, attacked_by, same_file
Conventions: pawns cover the two squares diagonally forward (white towards rank 8); rooks, bishops and queens are blocked by the first piece in their path; check, pins and the side to move are ignored.
\end{Verbatim}
\paragraph{Chess item, \cond{+rel.}: fact block}
\begin{Verbatim}[breaklines=true,breakanywhere=true,fontsize=\footnotesize,frame=single]
The following facts were computed by a deterministic script (CONFIRMED BY CODE - DO NOT RECALCULATE, USE AS TRUTH):

- Arriving piece: black rook on a8
- Relations between the arriving piece and each piece (only these pairs; relations among Pieces 1-4 are not listed):
  - Arriving piece x Piece 1 (white pawn on c2): none
  - Arriving piece x Piece 2 (black knight on c7): defended_by
  - Arriving piece x Piece 3 (black queen on h6): none
  - Arriving piece x Piece 4 (white queen on a5): attacks, attacked_by, same_file
\end{Verbatim}

\section{Decision rules for H1--H9}
\label{app:hypotheses}

The hypotheses were written before any model was queried:
\begin{itemize}
\item \textbf{H1.} Closed-book single look-ups (stem and branch element, hidden stems) are near ceiling; errors concentrate in memorised long tables (NaYin), set-valued relations (branch relations, mainly recall), arithmetic (day pillar) and multi-step chains (ten god, transit ten god).
\item \textbf{H2.} The reference table in context removes most look-up errors, but much less for the day pillar and the transit ten god, which require counting over the table.
\item \textbf{H3.} The \cond{tool} condition is at or near ceiling in every category; residual errors are misquotes.
\item \textbf{H4.} Errors fall with model size, but the closed-book day-pillar error stays large for every model.
\end{itemize}
The decision rules below were fixed after the open-weight runs had been inspected and before any confirmatory run; the thresholds are those of the original analysis plan.
\begin{itemize}
\item \textbf{H1.} ``Near ceiling'': closed-book accuracy $\geq 95\%$ in each of the three look-up categories. ``Errors concentrate'': the closed-book error of each of NaYin, branch relations, day pillar, ten god and transit ten god exceeds the pooled error of the look-up categories (one-sided Fisher exact test). ``Mainly recall'': branch-relation recall below precision.
\item \textbf{H2.} Look-up categories pooled (34 items): relative error reduction from \cond{closed} to \cond{table} $\geq 50\%$ with a significant McNemar test. Day pillar and transit ten god: relative reduction below the pooled look-up reduction.
\item \textbf{H3.} \cond{tool} accuracy $\geq 95\%$ in every category; each residual error is classified as a misquote or as a fact/gold mismatch.
\item \textbf{H4.} Macro error decreases Haiku $>$ Sonnet $>$ Opus on adjacent pairs (paired tests); closed-book day-pillar error $\geq 20\%$ for each model.
\end{itemize}

\paragraph{Collision set (H5--H9).}
These hypotheses were written and their decision rules locked on 2026-11-03, after the open-weight collision analysis of \S\ref{sec:collision} and before any confirmatory run. They add one condition, \cond{+full$-$natal}: the \cond{+full} output with the relation lines between two natal pillars removed and every other line unchanged. Against \cond{+full} it isolates the natal--natal relations; against \cond{+rel.} it isolates the remaining differences: \cond{+rel.} also omits the pillar attributes, the element interactions between pillars and the convention notes, and groups the relations by natal pillar with explicit ``none'' lines. The confirmatory runs pose all 160 collision items under \cond{closed}, \cond{table}, \cond{+pillar} (date variant), \cond{+attrs}, \cond{+full}, \cond{+full$-$natal} and \cond{+rel.} to the three proprietary models at effort \texttt{low}. The metric is exact accuracy over the four relation sets; a cell is one model and one variant (six cells); contrasts use the exact McNemar test with Holm correction over the six cells at $\alpha=0.05$; a cell in which both conditions reach 95\% is recorded as at ceiling and not decided. The reference rung is \cond{closed} in the pillar variant and \cond{+pillar} in the date variant.
\begin{itemize}
\item \textbf{H5.} Static attributes do not substitute for relations. A cell supports H5 unless \cond{+attrs} exceeds the reference rung by at least 10 points with a significant test; H5 holds if at least five of the decidable cells support it.
\item \textbf{H6.} Computed relations help: \cond{+full} $>$ \cond{+attrs} (one-sided). Supported if significant in every decidable cell, partly supported if in some.
\item \textbf{H7.} Filtering to the question helps: \cond{+rel.} $>$ \cond{+full} (one-sided), decided as H6.
\item \textbf{H8.} Source of the filtering gain. In cells where \cond{+rel.} exceeds \cond{+full} by at least 10 points, $R = (a_{\text{nonatal}} - a_{\text{full}}) / (a_{\text{rel}} - a_{\text{full}})$, where $a$ is exact accuracy under \cond{+full$-$natal}, \cond{+full} and \cond{+rel.}; $R \geq 0.5$ attributes the gain mainly to removing natal--natal relations, $R < 0.5$ mainly to the layout and the explicit ``none''. Without a qualifying cell, H8 is not decided on the confirmatory models.
\item \textbf{H9.} Under \cond{+full}, a wrong slot whose relation type the fact layer lists between that natal pillar and another natal pillar receives false positives more often than an unlisted slot (one-sided Fisher exact test per model, both variants pooled, Holm over the three models).
\end{itemize}

\paragraph{Chess set (C1--C4).}
Attacks and defences follow move coverage, with sliding pieces blocked by intervening pieces; the chess runs reuse each model's decoding settings and length caps from its collision run. Locked on 2026-11-05, before any chess request; the protocol and the hashes of items, requests and grading code carry an OpenTimestamps proof of that date. The metric is exact accuracy over the four relation sets; a cell is one model--mode run; contrasts use exact McNemar tests with Holm correction over the decidable cells at $\alpha=0.05$, and a cell in which both conditions reach 95\% is not decided.
\begin{itemize}
\item \textbf{C1.} Attributes do not substitute for relations: a cell supports C1 unless \cond{+attrs} exceeds \cond{closed} by at least 10 points with a significant two-sided test; C1 holds if at least 80\% of decidable cells support it.
\item \textbf{C2.} Filtering to the question helps: \cond{+rel.} $>$ \cond{+full} (one-sided); supported if at least 75\% of decidable cells are significant, partly supported if at least 50\%.
\item \textbf{C3.} The filtering gain does not come mainly from removing distractors: in cells where \cond{+rel.} exceeds \cond{+full} by at least 10 points, $R$ as in H8; supported if $R<0.5$ in at least two thirds of these cells.
\item \textbf{C4.} False positives follow listed distractors: under \cond{+full}, for each run with at least 30 false positives, wrong slots whose relation type the layer lists between that piece and another piece on the board (either direction) receive false positives more often than unlisted slots (one-sided Fisher exact test, Holm over runs); supported if significant in more than half of the qualifying runs.
\end{itemize}

\section{Deviations and exclusions}
\label{app:deviations}

\paragraph{Transit wording (item set v2).}
In v1, the transit question asked for ``公历 X 的日柱''. In the \cond{tool} condition the supplied facts label that date's pillar 流日 and use 日柱 for the natal day pillar, so the question pointed at the wrong line; most models copied the natal pillar and scored 0--2\% on this category under \cond{tool}. v2 asks for ``这一天的流日干支（当天的日柱，不是本命日柱）''. Item identifiers, dates, charts and gold answers are unchanged, and a unit test checks the wording. All transit rows of all runs were re-run with v2; v1 outputs are archived.

\paragraph{H2 pooling.}
The draft plan compared the day pillar and transit ten god with the median per-category look-up reduction and required each look-up category to be significant. The locked rule pools the 34 look-up items, because 10--12 items per category give too little power.

\paragraph{Holm scope in H5--H8.}
The locked rule corrects over the six cells of a contrast. Our first analysis script corrected over the decidable cells instead, which gave $p_{\text{Holm}}=0.032$ for H6 in all three decidable cells and would have made H6 supported. We found this in review after the runs, corrected the script, and report the locked correction here and in the main text. The decision on H5 does not change, and H7 and H8 have no decidable cell. The later chess protocol states a correction over decidable cells in its locked text, so C1--C4 are unaffected.

\paragraph{Decoding.}
No-thinking runs use greedy decoding; thinking runs use each model's released sampling configuration, because greedy decoding with thinking produced repetition loops in a pilot (14 of 108 outputs truncated for Qwen3-0.6B).

\paragraph{Excluded thinking runs.}
Gemma-4-31B-it never emitted its reasoning channel under schema-constrained decoding (0 of 2,046 requests); measuring it would require a two-stage decode. OLMo-3.1-32B-Think produced schema-conforming answers for 140 of 2,046 requests and reached the output cap on 400; in the collision set nearly all outputs were truncated. Raw outputs of both runs are archived with the release.

\paragraph{Additional experiments.}
The thinking-budget probe and the collision set were added after the first analysis of the open-weight runs, motivated by the day-pillar results; neither is part of H1--H4. After the first analysis of the collision set we added three further runs: the query-filtered condition \cond{+rel.} (all models), a 24k-token repeat of the most truncated collision conditions (three thinking models), and two replicate samples of selected thinking cells (four models). The original 720 collision requests were not changed. The \cond{+full$-$natal} condition of H8 was added on 2026-11-03, before any confirmatory run; its open-weight runs (all eleven models, same decoding settings as \cond{+rel.}) are exploratory and reported in Table~\ref{tab:h8}. The weekday control (\S\ref{sec:calendar}) was added after the main analysis and uses one prompt template.

\paragraph{Late chess runs.}
The two Qwen3.8-Flash-Next chess runs were made after the other 18 chess runs had been analysed, because the first job for them was removed from the cluster queue before it started. They use the locked requests, model revision and decoding settings. With them, C3 changes from supported (5 of 7 cells) to mixed (5 of 8); C1, C2 and C4 keep their decisions. Post hoc, and therefore exploratory, the eight C3 cells split by how well the model reads the complete list. Without thinking, $R$ is 0.75--0.88 for the three models that score 61--79\% under \cond{+full}, and between $-0.56$ and $0.29$ for the three that score 21--30\%; the two thinking cells (65--68\%) have $R \leq 0.10$. A model that already reads the complete list may lose most of its remaining items to attached distractors, while a weaker one also needs the shorter, regrouped layout. The split rests on eight cells and does not appear in the collision set, where Qwen3-32B has $R \leq 0.31$.

\paragraph{Confirmatory runs at effort \texttt{high}.}
The exploratory Opus runs at effort \texttt{high} ended in API errors for 388 of 398 closed-book and all 160 table requests and are not reported.

\section{Full results}
\label{app:full}

\begin{table*}[t]
\centering\scriptsize
\setlength{\tabcolsep}{3.5pt}
\begin{tabular}{llrrrrrrrrrrrrr}
\toprule
Model & Think & StE & BrE & ERel & TG & Hid & NaY & StR & BrR & Sea & DP & Abs & TrTG & Macro \\
\midrule
Majority & & 10 & 17 & 11 & 10 & 8 & 3 & 80 & 54 & 20 & 5 & 49 & 3 & 23 \\
\midrule
Qwen3-0.6B & -- & 80 & 50 & 11 & 10 & 0 & 2 & 0 & 0 & 20 & 0 & 18 & 0 & 16 \\
Qwen3-0.6B & \checkmark & 50 & 33 & 24 & 6 & 0 & 2 & 11 & 8 & 27 & 2 & 18 & 0 & 15 \\
Qwen3-1.7B & -- & 90 & 75 & 11 & 12 & 8 & 3 & 0 & 53 & 22 & 2 & 27 & 0 & 25 \\
Qwen3-1.7B & \checkmark & 80 & 75 & 22 & 8 & 0 & 3 & 53 & 37 & 43 & 2 & 52 & 0 & 31 \\
Qwen3-4B & -- & 100 & 75 & 27 & 21 & 8 & 7 & 56 & 0 & 27 & 3 & 51 & 0 & 31 \\
Qwen3-4B & \checkmark & 100 & 100 & 91 & 34 & 33 & 7 & 78 & 46 & 45 & 2 & 83 & 0 & 52 \\
Qwen3-8B & -- & 100 & 75 & 36 & 12 & 25 & 40 & 49 & 13 & 38 & 3 & 30 & 0 & 35 \\
Qwen3-8B & \checkmark & 100 & 100 & 93 & 44 & 33 & 45 & 89 & 59 & 52 & 1 & 91 & 0 & 59 \\
Qwen3-14B & -- & 100 & 92 & 47 & 21 & 42 & 78 & 80 & 10 & 40 & 0 & 71 & 2 & 49 \\
Qwen3-14B & \checkmark & 100 & 100 & 98 & 57 & 50 & 82 & 93 & 69 & 57 & 0 & 98 & 0 & 67 \\
Qwen3-32B & -- & 100 & 83 & 67 & 34 & 42 & 92 & 80 & 19 & 42 & 0 & 40 & 0 & 50 \\
Qwen3-32B & \checkmark & 100 & 100 & 93 & 45 & 42 & 87 & 96 & 78 & 65 & 2 & 99 & 0 & 67 \\
Qwen3.6-35B-A3B & -- & 100 & 92 & 56 & 100 & 100 & 100 & 89 & 67 & 55 & 0 & 63 & 0 & 68 \\
Qwen3.6-35B-A3B & \checkmark & 100 & 100 & 93 & 96 & 67 & 100 & 100 & 97 & 87 & 2 & 98 & 2 & 78 \\
Qwen3.8-27B & -- & 100 & 100 & 69 & 98 & 100 & 100 & 89 & 56 & 67 & 4 & 77 & 0 & 72 \\
Qwen3.8-27B & \checkmark & 100 & 100 & 89 & 91 & 92 & 98 & 100 & 97 & 85 & 86 & 100 & 57 & 91 \\
Qwen3.8-Flash-Next & -- & 100 & 100 & 93 & 100 & 100 & 100 & 91 & 73 & 85 & 4 & 68 & 0 & 76 \\
Qwen3.8-Flash-Next & \checkmark & 100 & 100 & 98 & 94 & 83 & 100 & 100 & 100 & 97 & 61 & 100 & 70 & 92 \\
Gemma-4-31B-it & -- & 100 & 100 & 98 & 100 & 92 & 97 & 96 & 68 & 67 & 3 & 88 & 0 & 76 \\
OLMo-3.1-32B-Inst. & -- & 40 & 17 & 24 & 9 & 0 & 0 & 60 & 8 & 20 & 1 & 26 & 0 & 17 \\
\bottomrule
\end{tabular}
\caption{Per-category accuracy (\%), \textsc{closed} condition. Abbreviations as in Table~\ref{tab:categories}.}
\label{tab:full-closed}
\end{table*}

\begin{table*}[t]
\centering\scriptsize
\setlength{\tabcolsep}{3.5pt}
\begin{tabular}{llrrrrrrrrrrrrr}
\toprule
Model & Think & StE & BrE & ERel & TG & Hid & NaY & StR & BrR & Sea & DP & Abs & TrTG & Macro \\
\midrule
Majority & & 10 & 17 & 11 & 10 & 8 & 3 & 80 & 54 & 20 & 5 & 49 & 3 & 23 \\
\midrule
Qwen3-0.6B & -- & 90 & 83 & 9 & 12 & 33 & 10 & 0 & 0 & 18 & 2 & 15 & 0 & 23 \\
Qwen3-0.6B & \checkmark & 80 & 75 & 22 & 7 & 58 & 8 & 4 & 4 & 23 & 2 & 27 & 0 & 26 \\
Qwen3-1.7B & -- & 100 & 92 & 13 & 16 & 58 & 27 & 67 & 1 & 25 & 2 & 35 & 0 & 36 \\
Qwen3-1.7B & \checkmark & 100 & 83 & 58 & 14 & 67 & 22 & 80 & 36 & 53 & 2 & 32 & 0 & 46 \\
Qwen3-4B & -- & 100 & 100 & 22 & 18 & 100 & 28 & 36 & 6 & 50 & 0 & 50 & 0 & 43 \\
Qwen3-4B & \checkmark & 100 & 100 & 98 & 59 & 92 & 80 & 100 & 78 & 100 & 9 & 97 & 2 & 76 \\
Qwen3-8B & -- & 100 & 100 & 36 & 15 & 92 & 50 & 40 & 13 & 48 & 2 & 46 & 0 & 45 \\
Qwen3-8B & \checkmark & 100 & 100 & 96 & 73 & 92 & 93 & 100 & 91 & 100 & 17 & 88 & 7 & 80 \\
Qwen3-14B & -- & 100 & 100 & 42 & 26 & 100 & 70 & 71 & 21 & 37 & 0 & 78 & 0 & 54 \\
Qwen3-14B & \checkmark & 100 & 100 & 100 & 92 & 75 & 100 & 100 & 95 & 100 & 14 & 100 & 2 & 81 \\
Qwen3-32B & -- & 100 & 100 & 42 & 31 & 100 & 60 & 87 & 24 & 43 & 1 & 61 & 0 & 54 \\
Qwen3-32B & \checkmark & 100 & 100 & 96 & 83 & 83 & 98 & 100 & 94 & 100 & 29 & 100 & 3 & 82 \\
Qwen3.6-35B-A3B & -- & 100 & 100 & 42 & 80 & 100 & 98 & 98 & 29 & 75 & 0 & 79 & 0 & 67 \\
Qwen3.6-35B-A3B & \checkmark & 100 & 100 & 96 & 92 & 83 & 100 & 98 & 99 & 100 & 45 & 100 & 50 & 89 \\
Qwen3.8-27B & -- & 100 & 100 & 58 & 96 & 100 & 100 & 84 & 74 & 88 & 2 & 65 & 0 & 72 \\
Qwen3.8-27B & \checkmark & 100 & 100 & 98 & 99 & 75 & 100 & 100 & 100 & 100 & 89 & 100 & 90 & 96 \\
Qwen3.8-Flash-Next & -- & 100 & 100 & 98 & 100 & 100 & 100 & 100 & 65 & 83 & 2 & 63 & 0 & 76 \\
Qwen3.8-Flash-Next & \checkmark & 100 & 100 & 100 & 99 & 100 & 100 & 100 & 99 & 100 & 95 & 100 & 90 & 99 \\
Gemma-4-31B-it & -- & 100 & 100 & 96 & 98 & 100 & 100 & 100 & 87 & 100 & 3 & 90 & 0 & 81 \\
OLMo-3.1-32B-Inst. & -- & 100 & 83 & 33 & 10 & 50 & 38 & 69 & 4 & 28 & 1 & 49 & 0 & 39 \\
\bottomrule
\end{tabular}
\caption{Per-category accuracy (\%), \textsc{table} condition. Abbreviations as in Table~\ref{tab:categories}.}
\label{tab:full-table}
\end{table*}

\begin{table*}[t]
\centering\scriptsize
\setlength{\tabcolsep}{3.5pt}
\begin{tabular}{llrrrrrrrrrrrrr}
\toprule
Model & Think & StE & BrE & ERel & TG & Hid & NaY & StR & BrR & Sea & DP & Abs & TrTG & Macro \\
\midrule
Majority & & 10 & 17 & 11 & 10 & 8 & 3 & 80 & 54 & 20 & 5 & 49 & 3 & 23 \\
\midrule
Qwen3-0.6B & -- & 100 & 92 & 18 & 99 & 92 & 52 & 40 & 35 & 28 & 93 & 6 & 3 & 55 \\
Qwen3-0.6B & \checkmark & 90 & 67 & 40 & 85 & 42 & 32 & 44 & 33 & 95 & 92 & 44 & 5 & 56 \\
Qwen3-1.7B & -- & 100 & 92 & 36 & 100 & 100 & 100 & 20 & 96 & 43 & 95 & 28 & 25 & 70 \\
Qwen3-1.7B & \checkmark & 100 & 100 & 71 & 100 & 75 & 100 & 20 & 65 & 50 & 83 & 28 & 18 & 68 \\
Qwen3-4B & -- & 100 & 83 & 78 & 100 & 100 & 100 & 100 & 100 & 98 & 100 & 58 & 98 & 93 \\
Qwen3-4B & \checkmark & 100 & 100 & 100 & 100 & 92 & 100 & 100 & 74 & 100 & 100 & 98 & 67 & 94 \\
Qwen3-8B & -- & 100 & 100 & 100 & 100 & 100 & 100 & 51 & 72 & 97 & 100 & 86 & 95 & 92 \\
Qwen3-8B & \checkmark & 100 & 92 & 100 & 99 & 75 & 100 & 100 & 85 & 100 & 100 & 95 & 73 & 93 \\
Qwen3-14B & -- & 100 & 100 & 100 & 100 & 100 & 100 & 100 & 100 & 100 & 100 & 98 & 97 & 100 \\
Qwen3-14B & \checkmark & 100 & 100 & 100 & 100 & 83 & 100 & 100 & 99 & 100 & 100 & 99 & 85 & 97 \\
Qwen3-32B & -- & 100 & 100 & 100 & 100 & 100 & 100 & 100 & 100 & 100 & 100 & 93 & 97 & 99 \\
Qwen3-32B & \checkmark & 100 & 100 & 100 & 100 & 67 & 100 & 100 & 100 & 100 & 100 & 100 & 93 & 97 \\
Qwen3.6-35B-A3B & -- & 100 & 100 & 100 & 100 & 100 & 100 & 100 & 100 & 100 & 100 & 100 & 100 & 100 \\
Qwen3.6-35B-A3B & \checkmark & 60 & 100 & 98 & 100 & 75 & 98 & 96 & 100 & 100 & 100 & 100 & 97 & 94 \\
Qwen3.8-27B & -- & 100 & 100 & 100 & 100 & 100 & 100 & 100 & 100 & 100 & 100 & 99 & 100 & 100 \\
Qwen3.8-27B & \checkmark & 100 & 100 & 98 & 100 & 92 & 100 & 100 & 100 & 100 & 100 & 100 & 100 & 99 \\
Qwen3.8-Flash-Next & -- & 100 & 100 & 100 & 100 & 100 & 100 & 100 & 100 & 100 & 100 & 98 & 100 & 100 \\
Qwen3.8-Flash-Next & \checkmark & 100 & 100 & 100 & 100 & 92 & 100 & 98 & 99 & 100 & 100 & 100 & 100 & 99 \\
Gemma-4-31B-it & -- & 100 & 100 & 100 & 100 & 100 & 100 & 100 & 100 & 100 & 100 & 100 & 100 & 100 \\
OLMo-3.1-32B-Inst. & -- & 100 & 100 & 100 & 100 & 75 & 97 & 100 & 100 & 43 & 99 & 82 & 5 & 83 \\
\bottomrule
\end{tabular}
\caption{Per-category accuracy (\%), \textsc{tool} condition. Abbreviations as in Table~\ref{tab:categories}.}
\label{tab:full-tool}
\end{table*}

\begin{table*}[t]
\centering\small
\setlength{\tabcolsep}{4pt}
\begin{tabular}{llrrrrrrrrrrr}
\toprule
 & & \multicolumn{5}{c}{Transit pillar given} & \multicolumn{6}{c}{Only the date given} \\
\cmidrule(lr){3-7}\cmidrule(lr){8-13}
Model & Think & closed & +attrs & +full & table & +rel. & closed & +pillar & +attrs & +full & table & +rel. \\
\midrule
Qwen3-0.6B & -- & 0 & 0 & 0 & 0 & 0 & 0 & 0 & 0 & 0 & 0 & 1 \\
Qwen3-1.7B & -- & 0 & 0 & 0 & 0 & 42 & 0 & 0 & 0 & 0 & 0 & 32 \\
Qwen3-4B & -- & 0 & 0 & 0 & 0 & 54 & 0 & 0 & 0 & 0 & 0 & 49 \\
Qwen3-8B & -- & 0 & 0 & 1 & 0 & 81 & 1 & 1 & 0 & 0 & 0 & 78 \\
Qwen3-14B & -- & 0 & 0 & 0 & 0 & 68 & 0 & 0 & 0 & 0 & 0 & 60 \\
Qwen3-32B & -- & 0 & 0 & 1 & 1 & 92 & 0 & 2 & 0 & 0 & 0 & 90 \\
Qwen3.6-35B-A3B & -- & 0 & 0 & 24 & 0 & 100 & 0 & 0 & 0 & 19 & 0 & 96 \\
Qwen3.8-27B & -- & 0 & 0 & 28 & 0 & 100 & 0 & 0 & 0 & 16 & 0 & 100 \\
Qwen3.8-Flash-Next & -- & 0 & 0 & 31 & 0 & 100 & 0 & 0 & 0 & 20 & 0 & 100 \\
Gemma-4-31B-it & -- & 0 & 1 & 68 & 0 & 100 & 0 & 2 & 1 & 55 & 0 & 100 \\
OLMo-3.1-32B-Inst. & -- & 0 & 0 & 0 & 0 & 85 & 0 & 0 & 0 & 0 & 0 & 38 \\
\midrule
Qwen3-0.6B & \checkmark & 0 & 0 & 0 & 0 & 1 & 0 & 0 & 0 & 0 & 0 & 1 \\
Qwen3-1.7B & \checkmark & 0 & 1 & 0 & 4 & 35 & 0 & 0 & 0 & 0 & 0 & 24 \\
Qwen3-4B & \checkmark & 1 & 4 & 8 & 12 & 54 & 0 & 0 & 2 & 2 & 0 & 62 \\
Qwen3-8B & \checkmark & 5 & 4 & 18 & 29 & 82 & 2 & 6 & 1 & 16 & 1 & 92 \\
Qwen3-14B & \checkmark & 5 & 8 & 57 & 49 & 96 & 1 & 9 & 4 & 49 & 2 & 94 \\
Qwen3-32B & \checkmark & 18 & 22 & 49 & 64 & 94 & 2 & 11 & 11 & 28 & 2 & 90 \\
Qwen3.6-35B-A3B & \checkmark & 74 & 64 & 90 & 95 & 99 & 1 & 65 & 72 & 85 & 28 & 95 \\
Qwen3.8-27B & \checkmark & 82 & 68 & 96 & 100 & 100 & 2 & 85 & 69 & 99 & 26 & 100 \\
Qwen3.8-Flash-Next & \checkmark & 91 & 79 & 98 & 91 & 100 & 9 & 92 & 88 & 94 & 62 & 99 \\
\bottomrule
\end{tabular}
\caption{Transit--natal collisions: share of items (\%, $n=80$ per cell) whose complete set of pairwise relations between the transit day and the four natal pillars is exactly right. \textit{+pillar}: the computed transit pillar is supplied; \textit{+attrs}: polarity, element and hidden stems of every pillar (including the transit pillar) are supplied; \textit{+full}: the fact layer's full output, including all relations; \textit{table}: the reference table instead of facts; \textit{+rel.}: only the transit pillar, its ten god and the relations between the transit day and each natal pillar (query-filtered fact layer).}
\label{tab:collision}
\end{table*}

\begin{table}[t]
\centering\scriptsize
\setlength{\tabcolsep}{2.5pt}
\begin{tabular}{llrrrrrr}
\toprule
Model & Th. & closed & table & +attrs & +full & $-$natal & +rel. \\
\midrule
Qwen3-0.6B & -- & 0 & 0 & 0 & 0 & 0 & 0 \\
Qwen3-1.7B & -- & 0 & 0 & 0 & 1 & 2 & 2 \\
Qwen3-4B & -- & 0 & 0 & 0 & 21 & 26 & 71 \\
Qwen3-8B & -- & 0 & 0 & 0 & 30 & 24 & 41 \\
Qwen3-14B & -- & 0 & 0 & 0 & 32 & 45 & 42 \\
Qwen3-32B & -- & 0 & 0 & 0 & 71 & 79 & 81 \\
Qwen3.6-35B-A3B & -- & 0 & 0 & 0 & 61 & 86 & 95 \\
Qwen3.8-27B & -- & 0 & 0 & 0 & 89 & 88 & 98 \\
Qwen3.8-Flash-Next & -- & 1 & 0 & 1 & 79 & 98 & 100 \\
Gemma-4-31B-it & -- & 1 & 1 & 6 & 91 & 96 & 100 \\
OLMo-3.1-32B-Inst. & -- & 0 & 0 & 0 & 24 & 35 & 62 \\
\midrule
Qwen3-0.6B & \checkmark & 0 & 0 & 0 & 1 & 1 & 3 \\
Qwen3-1.7B & \checkmark & 0 & 2 & 0 & 1 & 2 & 3 \\
Qwen3-4B & \checkmark & 22 & 22 & 17 & 68 & 70 & 94 \\
Qwen3-8B & \checkmark & 24 & 39 & 22 & 82 & 78 & 57 \\
Qwen3-14B & \checkmark & 42 & 58 & 36 & 65 & 62 & 97 \\
Qwen3-32B & \checkmark & 40 & 66 & 47 & 85 & 89 & 84 \\
Qwen3.6-35B-A3B & \checkmark & 80 & 92 & 80 & 94 & 96 & 100 \\
Qwen3.8-27B & \checkmark & 98 & 99 & 96 & 100 & 100 & 100 \\
Qwen3.8-Flash-Next & \checkmark & 99 & 100 & 95 & 99 & 99 & 100 \\
\midrule
Claude Haiku 5.5 & low & 73 & 85 & 74 & 98 & 99 & 98 \\
Claude Sonnet 5.5 & low & 96 & 97 & 99 & 99 & 100 & 100 \\
Claude Opus 5.5 & low & 97 & 98 & 100 & 100 & 100 & 100 \\
\bottomrule
\end{tabular}
\caption{Chess set: share of items (\%, $n=160$) whose relation sets between the new piece and all four pieces on the board are exactly right. Conditions mirror the collision ladder; $-$natal is \cond{+full} without the relations between two pieces already on the board. Th.: thinking; low: Claude at effort \texttt{low}.}
\label{tab:chess}
\end{table}

\begin{figure*}[t]
\centering
\includegraphics[width=\textwidth]{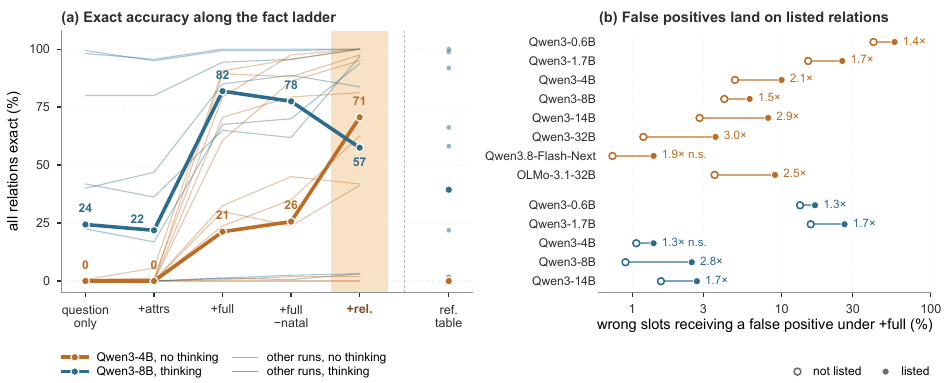}
\caption{Chess set in detail. (a) Exact accuracy along the fact ladder; the shaded column is \cond{+rel.}, and the reference table is shown apart because it is not a step of the ladder. Qwen3-4B without thinking gains most from filtering, and Qwen3-8B with thinking loses from it. Where the filtering gain is at least 10 points, removing only the relations between pieces already on the board recovers less than half of it, except for Qwen3-32B ($R=0.82$), Qwen3.6 ($R=0.75$) and Qwen3.8-Flash-Next ($R=0.88$) without thinking. (b) Under \cond{+full}, wrong slots whose relation type the list states receive more false positives than unlisted slots, for the 13 runs with at least 30 false positives. The axis is logarithmic, so connector length is the ratio. n.s.: not significant after Holm correction (one-sided Fisher exact test).}
\label{fig:chess-detail}
\end{figure*}

\begin{figure*}[t]
\centering
\includegraphics[width=\textwidth]{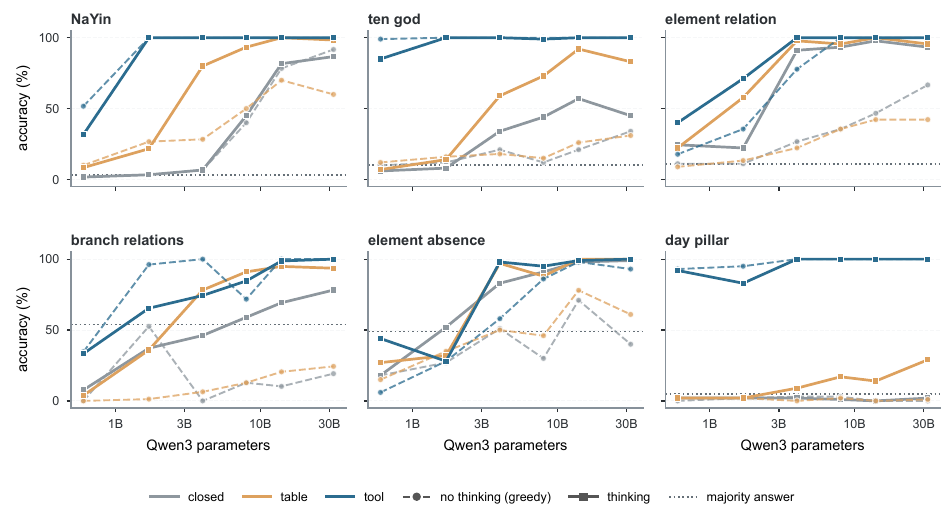}
\caption{Qwen3 0.6B--32B: accuracy against parameter count in six categories, by condition (colour) and mode (dashed: no thinking; solid: thinking). Dotted line: majority-class answer. Memorised-table categories rise with size; the closed-book day pillar stays near zero.}
\label{fig:scaling}
\end{figure*}

\section{Additional results}
\label{app:extra}

\begin{table}[t]
\centering\scriptsize
\setlength{\tabcolsep}{3pt}
\begin{tabular}{lllrrrrr}
\toprule
 & & & \multicolumn{2}{c}{8k cap} & \multicolumn{3}{c}{24k cap} \\
\cmidrule(lr){4-5}\cmidrule(lr){6-8}
Model & Cat. & Cond. & Acc. & Trunc. & Acc. & Trunc. & Med. tok. \\
\midrule
Qwen3-32B & DP & closed & 2 & 0 & 2 & 0 & 288 \\
 & DP & table & 29 & 2 & 35 & 0 & 1{,}926 \\
 & TrTG & closed & 0 & 0 & 0 & 0 & 695 \\
 & TrTG & table & 3 & 0 & 5 & 0 & 1{,}102 \\
\midrule
Qwen3.6-35B-A3B & DP & closed & 2 & 55 & 6 & 1 & 8{,}227 \\
 & DP & table & 45 & 42 & 69 & 1 & 7{,}713 \\
 & TrTG & closed & 2 & 21 & 3 & 0 & 6{,}436 \\
 & TrTG & table & 50 & 15 & 62 & 0 & 5{,}345 \\
\midrule
Qwen3.8-27B & DP & closed & 86 & 12 & 92 & 0 & 5{,}992 \\
 & DP & table & 89 & 9 & 98 & 0 & 4{,}810 \\
 & TrTG & closed & 57 & 23 & 85 & 0 & 6{,}622 \\
 & TrTG & table & 90 & 4 & 98 & 0 & 5{,}101 \\
\midrule
Qwen3.8-Flash-Next & DP & closed & 61 & 27 & 87 & 0 & 5{,}814 \\
 & DP & table & 95 & 1 & 96 & 0 & 2{,}612 \\
 & TrTG & closed & 70 & 15 & 97 & 0 & 5{,}901 \\
 & TrTG & table & 90 & 5 & 100 & 0 & 4{,}035 \\
\bottomrule
\end{tabular}
\caption{Thinking-budget probe. Accuracy (\%), number of truncated outputs, and median output tokens at the 24k cap. DP: 100 items; TrTG: 60 items.}
\label{tab:budget}
\end{table}

\paragraph{Thinking budget.}
The day pillar needs a count of up to about 55,000 days from a sexagenary anchor. At 24k, Qwen3-32B answers 2\% of closed-book day pillars with a median of 288 output tokens, too few for the count. Qwen3.6 reasons for a median of 8,227 tokens and reaches 6\% closed-book, but 69\% with the reference table, which contains the anchor 1949-10-01 = 甲子. Across three samples, the closed-book day pillar ranges over 73--86\% for Qwen3.8-27B and 61--88\% for Qwen3.8-Flash-Next. At the 24k cap, the closed-book transit ten god rises to 85\% for Qwen3.8-27B (57\% at 8k) and to 97\% for Qwen3.8-Flash-Next (70\% at 8k); Qwen3.6 is truncated on one closed-book day-pillar item (Table~\ref{tab:budget}).

\begin{table}[t]
\centering\scriptsize
\setlength{\tabcolsep}{3pt}
\begin{tabular}{llrrrrr}
\toprule
 & & \multicolumn{3}{c}{closed} & table & tool \\
\cmidrule(lr){3-5}\cmidrule(lr){6-6}\cmidrule(lr){7-7}
Model & Th. & WD & DP & $\pm$1 & WD & WD \\
\midrule
Qwen3-0.6B & -- & 9 & 0 & 33 & 15 & 100 \\
 & \checkmark & 11 & 2 & 35 & 13 & 100 \\
Qwen3-1.7B & -- & 20 & 2 & 35 & 11 & 100 \\
 & \checkmark & 22 & 2 & 32 & 15 & 100 \\
Qwen3-4B & -- & 34 & 3 & 67 & 51 & 100 \\
 & \checkmark & 62 & 2 & 86 & 40 & 100 \\
Qwen3-8B & -- & 49 & 3 & 76 & 58 & 100 \\
 & \checkmark & 75 & 1 & 80 & 51 & 100 \\
Qwen3-14B & -- & 57 & 0 & 63 & 54 & 100 \\
 & \checkmark & 82 & 0 & 88 & 68 & 100 \\
Qwen3-32B & -- & 45 & 0 & 82 & 41 & 100 \\
 & \checkmark & 83 & 2 & 88 & 49 & 100 \\
Qwen3.6-35B-A3B & -- & 69 & 0 & 77 & 68 & 100 \\
 & \checkmark & 93 & 2 & 75 & 89 & 100 \\
Qwen3.8-27B & -- & 57 & 4 & 84 & 63 & 100 \\
 & \checkmark & 97 & 86 & 50 & 93 & 100 \\
Qwen3.8-Flash-Next & -- & 84 & 4 & 75 & 74 & 100 \\
 & \checkmark & 97 & 61 & 100 & 96 & 100 \\
Gemma-4-31B-it & -- & 73 & 3 & 96 & 75 & 100 \\
OLMo-3.1-32B-Inst. & -- & 60 & 1 & 62 & 53 & 100 \\
\bottomrule
\end{tabular}
\caption{Weekday control on the 100 day-pillar dates. WD: weekday accuracy (\%; majority answer 21\%, uniform 14\%); DP: closed-book day-pillar accuracy on the same dates from the main runs; $\pm$1: share of wrong weekdays that are one day off (uniform 33\%). Th.: thinking.}
\label{tab:weekday}
\end{table}

\paragraph{Weekday control.}
With thinking, Qwen3-4B to Qwen3-32B and Qwen3.6 answer 62--93\% of weekdays against 0--2\% of day pillars. The reference table for this control has a weekday anchor line added. Supplying the sexagenary anchor does not close the gap: with the reference table, thinking Qwen3-32B answers 29\% of day pillars and Qwen3.6 45\%, against 83\% and 93\% of closed-book weekdays. With thinking, the two Qwen3.8 models answer 97\% of weekdays. Qwen3-0.6B and Qwen3-1.7B are near chance on both tasks (9--22\% of weekdays), and their wrong weekdays are one day off at the uniform rate (32--35\%). The reference table has mixed effects on the weekday (Qwen3-4B without thinking 34$\to$51\%; Qwen3-32B with thinking 83$\to$49\%, with 13 truncated outputs), and every model copies a computed weekday (100\%).

\begin{table*}[t]
\centering\small
\setlength{\tabcolsep}{4pt}
\begin{tabular}{lrrrrrrrrr}
\toprule
 & \multicolumn{6}{c}{Day pillar, closed-book} & table & \multicolumn{2}{c}{Weekday, closed-book} \\
\cmidrule(lr){2-7}\cmidrule(lr){8-8}\cmidrule(lr){9-10}
Model & Acc. & corr.\ anchor & corr.\ count & no anchor & wrong only & trunc. & corr.\ count & anchor prec. & formula \\
\midrule
Qwen3-4B & 2 & 1 & 0 & 77 & 17 & 3 & 26 & 63 & 99 \\
Qwen3-8B & 1 & 0 & 0 & 71 & 25 & 3 & 51 & 75 & 97 \\
Qwen3-14B & 0 & 0 & 0 & 72 & 27 & 1 & 36 & 72 & 93 \\
Qwen3-32B & 2 & 0 & 0 & 82 & 16 & 0 & 39 & 77 & 99 \\
Qwen3.6-35B-A3B & 2 & 61 & 47 & 0 & 20 & 55 & 93 & 72 & 99 \\
Qwen3.8-27B & 86 & 100 & 97 & 0 & 0 & 12 & 92 & 82 & 92 \\
Qwen3.8-Flash-Next & 61 & 99 & 94 & 1 & 0 & 27 & 99 & 90 & 90 \\
\bottomrule
\end{tabular}
\caption{Thinking traces, graded step by step (\%, 100 items per cell, 8k cap). \textit{corr.\ anchor}: the trace states at least one correct date--day-pillar pair other than the target; \textit{corr.\ count}: it contains the correct day count from such a pair, or the target's Julian Day Number; \textit{no anchor}, \textit{wrong only}, \textit{trunc.}: share of traces that state no date--pillar pair, only wrong ones, or end at the cap without an answer. Weekday: share of stated date--weekday pairs that are correct, and share of traces that name a weekday formula (Zeller, Kim Larsen or Doomsday).}
\label{tab:cot}
\end{table*}

\paragraph{Thinking traces.}
We graded the thinking traces of the day-pillar runs step by step (exploratory). A date--pillar statement is a Gregorian date other than the target followed by a stem--branch pair that is not marked as a year or month pillar, graded against the true day pillar of that date; a count is correct if the trace contains the day difference between the target and a correctly stated date, or the target's Julian Day Number. A random sample of 25 extracted statements contained no extraction error. With the reference table, which states the anchor, the Qwen3 models use it in 99--100\% of traces and the failure moves to the count: 26--51\% of traces contain the correct count, and among those that do not, half of the closest stated counts of Qwen3-8B and Qwen3-32B are within two days of the true one (median one day). Qwen3.6 reaches the 8k cap in 55\% of closed-book traces; at 24k it reaches it in 1\% and still answers 6\% correctly (Table~\ref{tab:budget}), with 12\% of its date--pillar statements correct and a year pillar used as a day pillar in 88\% of its traces. At 24k, 39\% of the wrong answers of Qwen3.6 and 67\% of those of Qwen3.8-Flash-Next are the pillar that a stated wrong pair implies under a correct count, against 6\% and 3\% expected by chance. Traces need not reflect the computation that produced an answer \citep{turpin2023unfaithful}; the analysis shows which visible steps accompany correct answers.

\paragraph{Relation set and binding under \cond{+full}.}
To locate what thinking changes in \cond{+full} answers, we also score each answer as set-exact, binding error and broadcast (definitions in Appendix~\ref{app:formal}; exploratory). Without thinking, Qwen3.6, Qwen3.8-27B and Qwen3.8-Flash-Next are set-exact on 34--46\% of BaZi \cond{+full} items and broadcast a relation in 36--42\%, against 98--100\% exact under \cond{+rel.}; 68--75\% of their false positives fall on natal pairs listed in the context, which make up 22\% of candidate slots. With thinking they are set-exact on 88--99\% and broadcast in 1--4\%. Thinking raises set-exact for Qwen3-8B to Qwen3-32B from 6--13\% to 34--65\%, while binding errors remain in 12--22\% and broadcasts in 11--34\% of their items. In chess, thinking lowers broadcasts of Qwen3-4B to Qwen3-14B from 30--48\% to 10--18\%, and to 0\% for Qwen3.6 and Qwen3.8. Among the remaining errors of Qwen3-8B to Qwen3-32B under thinking, 21--35\% have the correct relation set in the wrong pillars and 23--55\% contain a broadcast.

\begin{table*}[t]
\centering\small
\setlength{\tabcolsep}{3.5pt}
\begin{tabular}{lrrrrrrrrrr}
\toprule
 & \multicolumn{2}{c}{P: closed} & \multicolumn{2}{c}{P: +attrs} & \multicolumn{2}{c}{D: closed} & \multicolumn{2}{c}{D: table} & \multicolumn{2}{c}{D: +attrs} \\
 & 8k & 24k & 8k & 24k & 8k & 24k & 8k & 24k & 8k & 24k \\
\midrule
Qwen3.6-35B-A3B & 74 (0) & 69 (0) & 64 (0) & 75 (0) & 1 (43) & 1 (1) & 28 (37) & 46 (0) & 72 (0) & 62 (0) \\
Qwen3.8-27B & 82 (13) & 96 (0) & 68 (17) & 94 (0) & 2 (79) & 89 (4) & 26 (58) & 100 (0) & 69 (20) & 95 (0) \\
Qwen3.8-Flash-Next & 91 (3) & 96 (0) & 79 (12) & 94 (0) & 9 (75) & 94 (0) & 62 (31) & 99 (0) & 88 (7) & 94 (0) \\
\bottomrule
\end{tabular}
\caption{Collision accuracy (\%, all four relation sets exact) with thinking at an 8k and a 24k output cap; number of truncated outputs (of 80) in parentheses. P: transit pillar given; D: only the date given.}
\label{tab:collision-budget}
\end{table*}

\paragraph{Collisions with thinking: the date is the bottleneck.}
In the date variant at the 8k cap, the strongest thinking models score 1--9\% closed-book with most outputs truncated; at 24k, Qwen3.8-27B reaches 89\% and Qwen3.8-Flash-Next 94\%, and the transit pillar alone gives 85\% and 92\% at 8k. Qwen3.6 stays at 1\% at 24k, but 69\% of its relation sets are exactly right relative to its own wrong transit pillar: its collision errors come from the date.

\paragraph{Collisions.}
At the 24k cap, both Qwen3.8 models score 96\% closed-book and 94\% with attributes in the pillar variant. In the date variant they score 95\% (Qwen3.8-27B) and 94\% (Qwen3.8-Flash-Next) with attributes, and 100\% and 99\% with the reference table; Qwen3.8-27B has 4 truncated closed-book outputs. With thinking, \cond{+rel.} shortens reasoning relative to \cond{+full}: Qwen3.8-Flash-Next uses 837 instead of 3,168 output tokens in the pillar variant. For Qwen3.6, which was not truncated at either cap, the attribute effect in the pillar variant changes sign between the two runs ($74\to64\%$ at 8k, $69\to75\%$ at 24k). Under \cond{+full}, Gemma-4 without thinking reaches 68\% and 55\% (pillar and date variants); a wrong slot whose relation type the list states between that natal pillar and another natal pillar receives a false positive at 24.0\% against 2.6\% for unlisted slots in Qwen3.6 and at 18.5\% against 2.5\% in Qwen3.8-Flash-Next without thinking, whereas in the 0.6B--4B models the two rates are within a factor of 1.4, because those models assign relations to almost every slot. Under \cond{+rel.} without thinking, OLMo-3.1-32B-Instruct reaches 85\% and 38\% and Qwen3-1.7B 42\% and 32\%. With thinking, Qwen3.8-27B uses 2,182 instead of 4,584 output tokens on average in the pillar variant and 1,568 instead of 4,447 in the date variant, and Qwen3.8-Flash-Next 646 instead of 2,433 in the date variant. In the date variant at 24k, Qwen3.8-27B and Qwen3.8-Flash-Next produce means of 15,095 and 12,907 output tokens; supplying the transit pillar gives Qwen3.6 65\% at 8k. The Qwen3.6 share of exactly right relation sets among items with a wrong transit pillar is 69\% in \cond{closed} ($n=36$) and 78\% in \cond{table} ($n=18$); for Qwen3 models with thinking it is at most 38\%, so both steps fail.

\begin{table}[t]
\centering\scriptsize
\setlength{\tabcolsep}{2.5pt}
\begin{tabular}{llrrrrrrrr}
\toprule
 & & \multicolumn{4}{c}{Transit pillar given} & \multicolumn{4}{c}{Only the date given} \\
\cmidrule(lr){3-6}\cmidrule(lr){7-10}
Model & Th. & +full & $-$natal & +rel. & $R$ & +full & $-$natal & +rel. & $R$ \\
\midrule
Qwen3-1.7B & -- & 0 & 0 & 42 & 0.00 & 0 & 0 & 32 & 0.00 \\
Qwen3-4B & -- & 0 & 0 & 54 & 0.00 & 0 & 1 & 49 & 0.03 \\
Qwen3-8B & -- & 1 & 2 & 81 & 0.02 & 0 & 4 & 78 & 0.05 \\
Qwen3-14B & -- & 0 & 5 & 68 & 0.07 & 0 & 6 & 60 & 0.10 \\
Qwen3-32B & -- & 1 & 10 & 92 & 0.10 & 0 & 1 & 90 & 0.01 \\
Qwen3.6-35B-A3B & -- & 24 & 56 & 100 & 0.43 & 19 & 35 & 96 & 0.21 \\
Qwen3.8-27B & -- & 28 & 79 & 100 & 0.71 & 16 & 59 & 100 & 0.51 \\
Qwen3.8-Flash-Next & -- & 31 & 57 & 100 & 0.38 & 20 & 55 & 100 & 0.44 \\
Gemma-4-31B-it & -- & 68 & 74 & 100 & 0.19 & 55 & 61 & 100 & 0.14 \\
OLMo-3.1-32B-Inst. & -- & 0 & 0 & 85 & 0.00 & 0 & 0 & 38 & 0.00 \\
\midrule
Qwen3-1.7B & \checkmark & 0 & 0 & 35 & 0.00 & 0 & 0 & 24 & 0.00 \\
Qwen3-4B & \checkmark & 8 & 5 & 54 & -0.05 & 2 & 6 & 62 & 0.06 \\
Qwen3-8B & \checkmark & 18 & 16 & 82 & -0.02 & 16 & 29 & 92 & 0.16 \\
Qwen3-14B & \checkmark & 57 & 57 & 96 & 0.00 & 49 & 66 & 94 & 0.39 \\
Qwen3-32B & \checkmark & 49 & 62 & 94 & 0.31 & 28 & 40 & 90 & 0.20 \\
Qwen3.6-35B-A3B & \checkmark & 90 & 95 & 99 & -- & 85 & 95 & 95 & 1.00 \\
Qwen3.8-27B & \checkmark & 96 & 100 & 100 & -- & 99 & 95 & 100 & -- \\
Qwen3.8-Flash-Next & \checkmark & 98 & 89 & 100 & -- & 94 & 95 & 99 & -- \\
\bottomrule
\end{tabular}
\caption{Ablation of the filtering gain (H8, exploratory). Exact collision accuracy (\%, $n=80$) under \cond{+full}, \cond{+full$-$natal} (relations between two natal pillars removed) and \cond{+rel.}; $R=(a_{-\text{natal}}-a_{\text{full}})/(a_{\text{rel}}-a_{\text{full}})$, computed where \cond{+rel.} exceeds \cond{+full} by at least 10 points. Qwen3-0.6B has no such cell and is omitted.}
\label{tab:h8}
\end{table}

\paragraph{Filtering ablation (H8).}
\cond{+full$-$natal} removes only the relation lines between two natal pillars from \cond{+full}; \cond{+rel.} further omits the pillar attributes, the element interactions and the convention notes, groups the relations by natal pillar with explicit ``none'' lines, and is a quarter as long (428 against 1,758 characters). Length, per-pillar layout and ``none'' lines change together, so these runs do not separate them, and $R$ is a residual of this one ablation order. Table~\ref{tab:h8} reports $R$ in the 31 open-weight cells where \cond{+rel.} exceeds \cond{+full} by at least 10 points. $R$ is at most 0.06 for OLMo, Qwen3-1.7B and Qwen3-4B, 0.14--0.19 for Gemma-4, 0.38--0.44 for Qwen3.8-Flash-Next without thinking and 0.51--0.71 for Qwen3.8-27B without thinking. Qwen3-32B without thinking reaches 1\% under \cond{+full$-$natal} in the date variant, against 90\% under \cond{+rel.} Without thinking, removing the natal--natal lines lowers false positives per item in every model from 8B upward (Qwen3-32B: 5.74$\to$4.48 in the date variant), but not in Qwen3-1.7B and Qwen3-4B, which keep 13--22 false positives per item. Thinking Qwen3.8-Flash-Next falls from 98\% under \cond{+full} to 89\% under \cond{+full$-$natal} in the pillar variant ($p=0.039$).

\begin{table}[t]
\centering\small
\setlength{\tabcolsep}{4pt}
\begin{tabular}{lrrrr}
\toprule
Model & Err. & $\bar H$ & AUROC & Conf.\ err. \\
\midrule
Qwen3-0.6B & 91 & 1.15 & 0.62 & 17 \\
Qwen3-1.7B & 82 & 1.08 & 0.63 & 26 \\
Qwen3-4B & 78 & 0.87 & 0.76 & 34 \\
Qwen3-8B & 76 & 0.93 & 0.70 & 24 \\
Qwen3-14B & 63 & 0.66 & 0.79 & 32 \\
Qwen3-32B & 62 & 0.92 & 0.82 & 24 \\
Qwen3.6-35B-A3B & 40 & 1.09 & 0.89 & 8 \\
Qwen3.8-27B & 37 & 1.08 & 0.91 & 6 \\
Qwen3.8-Flash-Next & 33 & 0.92 & 0.93 & 8 \\
Gemma-4-31B-it & 32 & 0.61 & 0.86 & 27 \\
OLMo-3.1-32B-Inst. & 86 & 1.76 & 0.81 & 7 \\
\bottomrule
\end{tabular}
\caption{Sampling consistency, closed condition, no thinking (10 samples at $T=0.7$ per item). Err.: greedy error rate (\%); $\bar H$: mean answer entropy (bits); AUROC: entropy as a detector of greedy errors; Conf.\ err.: share of greedy errors on which all 10 samples agree (\%).}
\label{tab:e7}
\end{table}

\paragraph{Error signatures.}
Without thinking, 76--92\% of wrong day pillars have both stem and branch wrong (84.7\% under uniform guessing). Relations are over-reported: on closed-book branch relations without thinking, Qwen3 models that produce relations have a micro precision of 8--19\%, below their recall in every case (30--91\%). Thinking raises precision sharply (Qwen3-14B 19$\to$70\%, Qwen3-32B 17$\to$77\%, Qwen3.6 63$\to$100\%); Qwen3-1.7B without thinking returns almost only empty sets (4 predicted labels in 78 items). The pre-specified expectation that branch-relation errors are ``mainly recall'' holds in 3 of 20 runs. Table~\ref{tab:e7} reports ten samples per closed-book item without thinking: answer entropy separates greedy errors from correct answers with AUROC 0.62--0.93, but 6--34\% of greedy errors are reproduced by all ten samples, which a consistency-based detector \citep{manakul2023selfcheckgpt,farquhar2024semantic} cannot flag but a table check can.

\begin{table}[t]
\centering\small
\setlength{\tabcolsep}{4pt}
\begin{tabular}{llr}
\toprule
 & Locked decision rule & Runs \\
\midrule
H1 & closed look-ups $\geq 95\%$ (StE, BrE, Hid) & 2/20 \\
H1 & BrR recall $<$ precision & 3/20 \\
H2 & rule 1: pooled look-up error cut $\geq 50\%$, McNemar sig. & 10/18 \\
H2 & rule 2: DP and TrTG cut below look-up cut & 17/18 \\
H3 & tool $\geq 95\%$ in every category & 5/20 \\
H4 & closed DP error $\geq 20\%$ (no thinking) & 11/11 \\
H4 & closed DP error $\geq 20\%$ (thinking) & 8/9 \\
\bottomrule
\end{tabular}
\caption{Locked decision rules applied to the open-weight runs (exploratory; runs where the rule is satisfied / runs where it is decidable). H2 is undecidable when the closed look-up error is zero.}
\label{tab:hyp-open}
\end{table}

\paragraph{Pre-specified hypotheses on the open-weight runs.}
Table~\ref{tab:hyp-open} applies the locked decision rules to the 20 open-weight runs. H1's claim that single look-ups are at ceiling holds in 2 of 20 runs. Its claim that errors concentrate in five composite categories holds category by category: the closed-book error exceeds the pooled look-up error (one-sided Fisher test) for the day pillar in 19 of 20 runs and for the transit ten god in 20 of 20, but for ten gods, branch relations and NaYin only in 12, 11 and 9 of 20 runs, the categories that newer and larger models answer correctly without help. H2's second rule (the table helps the day pillar and transit ten god less than look-ups) holds in 17 of 18 decidable runs; its first rule holds in 10 of 18, the remaining runs mostly having too few closed-book look-up errors left to reach significance. H3 (\cond{tool} $\geq 95\%$ in every category) holds in 5 of 20 runs, all of them larger no-thinking models. H4's prediction that the closed-book day-pillar error stays at or above 20\% holds for all no-thinking runs and fails for Qwen3.8-27B with thinking (14\%).

\section{Confirmatory hypothesis tests}
\label{app:confirmatory}

The models were queried through the Claude Code CLI (version 2.1.317), with all tools disabled, no user or project settings loaded and our system prompt in place of the CLI's; every request returned an answer after at most one retry. The decision rules are those of Appendix~\ref{app:hypotheses}. Two choices that the locked text leaves open were fixed before the tests were computed: H4 is tested on the closed-book condition with two-sided exact McNemar tests (Holm over the two adjacent pairs), and H5 uses a two-sided test. Among the open-weight runs, the thinking runs are the closer comparison, because the confirmatory models reason at effort \texttt{low}.

\paragraph{H1--H4.}
Haiku's errors concentrate in the categories that combine attributes (day pillar, ten gods, transit ten god). The three look-up categories are at 100\% except for Opus on hidden stems (10 of 12 items), so H1's ceiling clause holds for Haiku and Sonnet. Errors concentrate significantly in the day pillar, ten gods and transit ten god for Haiku (three of the five categories) and in none for Sonnet and Opus, and branch-relation recall is not below precision for any model (Haiku: 97.7\% recall, 95.5\% precision). H2 cannot be decided for Haiku and Sonnet, whose look-up error is already 0\% closed-book; for Opus the look-up error falls from 5.9\% to 0\% ($p=0.50$), and neither H2 rule is met. Under H3, the 11 residual \cond{tool} errors are Haiku 8, Sonnet 1 and Opus 2. Closed-book error falls from Haiku to Sonnet (13.0\% to 3.1\% of items; $p_{\text{Holm}}<10^{-12}$) and from Sonnet to Opus (3.1\% to 1.3\%; $p_{\text{Holm}}=0.029$); the closed-book day-pillar error is 29\%, 4\% and 2\%.

\paragraph{H5--H9.}
A cell in which both conditions reach 95\% is at ceiling and not decided. H5 is not decided: its three decidable cells all support it (differences of $-5$ to $+5$ points, all $p_{\text{Holm}}=1$), but its threshold of five supporting cells cannot be reached because the other three cells are at ceiling. H6 is not supported: \cond{+full} exceeds \cond{+attrs} in Haiku in both variants (7.5 and 10 points; $p_{\text{Holm}}=0.078$ and $0.064$) and Opus in the date variant (6.25 points; $p_{\text{Holm}}=0.125$), none significant. H7 and H8 have no decidable cell, because \cond{+full} is already at 99--100\%; \cond{+rel.} and \cond{+full} differ by at most 1.25 points in either direction. Under H9, $p_{\text{Holm}}\geq0.146$ for every model.

\paragraph{Effort \texttt{high} (exploratory).}
With effort \texttt{high}, Haiku's closed-book day-pillar accuracy rises from 71\% to 95\% and its transit ten god from 62\% to 73\%; Sonnet reaches 96\% and 97\%. The Opus runs at effort \texttt{high} are not reported (Appendix~\ref{app:deviations}).

\section{Formal model of the fact layer}
\label{app:formal}

\paragraph{Exhaustive verification.}
Both rule sets are re-encoded as answer-set programs for clingo 5.8 \citep{gebser2019clingo}, without the fact-layer code or its table. Relations are arithmetic on the stem cycle $x,y \in \{0,\dots,9\}$ and the branch cycle $x,y \in \{0,\dots,11\}$: 天干五合 iff $|x-y|=5$, 天干相冲 iff $|x-y|=6$, 六合 iff $x+y \equiv 1$, 六害 iff $x+y \equiv 7$, 六冲 iff $|x-y|=6$, 六破 iff $\{x,y\}=\{b, b+9\}$ for an even $b$, and 半合 iff $x \neq y$, $x \equiv y \pmod 4$ and one of them is $\equiv 0 \pmod 3$ (all mod 12); 相刑 and 自刑 are enumerated. The day pillar is $(J+49) \bmod 60$ for the Julian Day Number $J$ \citep{fliegel1968machine}, not a count from an anchor date. Chess coverage casts rays that stop at the first occupied square. Each check in Table~\ref{tab:verify} covers every case of its finite domain, and all agree with the fact layers. To show that the checks are not vacuous, each of twelve mutants changes one rule (for example $b+9 \to b+3$ in 六破, no 400-year leap rule, no blocking), and every mutant fails at least one check \citep{jia2011mutation}.

\begin{table}[t]
\centering\small
\setlength{\tabcolsep}{3pt}
\begin{tabular}{p{0.66\columnwidth}rr}
\toprule
Check & Cases & Result \\
\midrule
B1 stem relations, all ordered pairs & 100 & pass \\
B2 branch relations, all ordered pairs & 144 & pass \\
B3 day pillar, every date 1900-01-01--2100-12-31 & 73{,}414 & pass \\
B4 ten god, all (day master, stem) pairs & 100 & pass \\
B5 collision items: gold relations, ten god, day pillar & 160 & pass \\
C1 coverage, every kind $\times$ colour $\times$ square (empty board) & 768 & pass \\
C2 sliding coverage, every rook/bishop/queen square $\times$ blocker & 12{,}096 & pass \\
C1+C2 fact-layer coverage vs python-chess & 12{,}864 & pass \\
C3 chess items: gold relations & 160 & pass \\
\bottomrule
\end{tabular}
\caption{Exhaustive checks of the fact layers against answer-set programs (B: BaZi, C: chess). Each check covers every case of its domain. Mutation test: 12 of 12 single-rule mutants fail at least one check.}
\label{tab:verify}
\end{table}

\paragraph{Computation graphs.}
Following \citet{dziri2023faith}, an item is the directed acyclic graph of the ground rule instances on which its scored answer, the four relation sets, depends. A condition states some nodes in the context. The residual graph contains the nodes reached backwards from the answer without passing a stated node, that is, the rule applications left to the model; $R$ is its size. Under \cond{+full} the per-slot sets are not stated, so each slot node selects from every relation line. Table~\ref{tab:formal} reports the residual graph of each condition. In BaZi, relations are defined on stem and branch identity, so element and polarity lie on no path to the answer and \cond{+attrs} leaves $R$ at 41. In chess, \cond{+attrs} states the coordinates and removes 10 nodes ($R$ 57.8 to 47.8), and accuracy with thinking stays at 56--58\%. Within each set, accuracy falls in three tiers ordered by $R$: about 41--58 nodes (relations to compute), 5 (selection from a list) and 1 (copying per slot). \cond{+full} and \cond{+full$-$natal} have the same residual graph and differ only in context lines that feed no residual node.

\begin{table}[t]
\centering\small
\setlength{\tabcolsep}{4pt}
\begin{tabular}{lrrrrrr}
\toprule
 & \multicolumn{2}{c}{Residual} & \multicolumn{2}{c}{Context lines} & \multicolumn{2}{c}{Exact (\%)} \\
\cmidrule(lr){2-3}\cmidrule(lr){4-5}\cmidrule(lr){6-7}
Condition & $R$ & depth & on & off & direct & thinking \\
\midrule
\multicolumn{7}{l}{\textit{BaZi collisions}} \\
\cond{closed} & 43.0 & 4.5 & 0.0 & 0.0 & 0.1 & 21.1 \\
\cond{table} & 43.0 & 4.5 & 0.0 & 0.0 & 0.1 & 40.2 \\
\cond{+pillar} & 41.0 & 3.0 & 1.0 & 0.0 & 0.7 & 38.4 \\
\cond{+attrs} & 41.0 & 3.0 & 1.0 & 4.0 & 0.1 & 35.4 \\
\cond{+full} & 5.0 & 2.0 & 3.1 & 34.3 & 14.6 & 56.2 \\
\cond{+full$-$natal} & 5.0 & 2.0 & 3.1 & 29.6 & 28.1 & 60.8 \\
\cond{+rel.} & 1.0 & 1.0 & 4.0 & 3.0 & 82.8 & 89.8 \\
\multicolumn{7}{l}{\textit{Chess}} \\
\cond{closed} & 57.8 & 6.0 & 0.0 & 0.0 & 0.1 & 58.0 \\
\cond{table} & 57.8 & 6.0 & 0.0 & 0.0 & 0.1 & 68.0 \\
\cond{+attrs} & 47.8 & 5.0 & 5.0 & 0.0 & 0.8 & 56.1 \\
\cond{+full} & 5.0 & 2.0 & 2.0 & 10.4 & 55.3 & 84.7 \\
\cond{+full$-$natal} & 5.0 & 2.0 & 2.0 & 8.0 & 64.0 & 84.7 \\
\cond{+rel.} & 1.0 & 1.0 & 4.0 & 2.0 & 76.7 & 90.3 \\
\bottomrule
\end{tabular}
\caption{Residual computation left by each condition (means over items) and exact accuracy (mean over open-weight models of at least 4B parameters, 8k cap). $R$: rule applications the model must perform; \textit{on}/\textit{off}: context lines that do or do not feed them.}
\label{tab:formal}
\end{table}

\paragraph{Item-level tests.}
Within a condition, the number of residual nodes varies across chess items with the blocking checks (SD 3.6), and the number of relation lines that do not involve the transit day or the arriving piece varies under \cond{+full} (SD 1.9 in BaZi, 1.0 in chess). Neither is associated with exact accuracy under thinking: odds ratio 0.99 [0.94, 1.04] per residual node, and 0.93 [0.84, 1.02] (BaZi) and 1.05 [0.87, 1.27] (chess) per such line, from a logit with one intercept per model, mode and condition and standard errors clustered by item; models answering directly give the same result. Date items, whose residual graph contains the calendar chain (depth 6 against 3), have an odds ratio of 0.018 [0.010, 0.032] against pillar items under thinking (\cond{closed} and \cond{table}). Accuracy follows which kind of node remains, not how many.

\paragraph{Error types.}
Let $S$ be the slots (pillars or pieces) and $T$ the relation types. An answer is a relation $A \subseteq S \times T$ with gold $G$, and $A_t = \{s : (s,t) \in A\}$. The answer is exact if $A = G$ and set-exact if $\pi_T(A) = \pi_T(G)$. A binding error is set-exact but not exact. A broadcast is a type $t \in \pi_T(G)$ with $|A_t| > |G_t|$.

\section{Reproducibility}
\label{app:repro}

\begin{table*}[t]
\centering\small
\begin{tabular}{llrl}
\toprule
Hugging Face repository & Revision & GPUs & Runs \\
\midrule
\texttt{Qwen/Qwen3-0.6B} & \texttt{c1899de289} & 1 & no-think, sampled, think \\
\texttt{Qwen/Qwen3-1.7B} & \texttt{70d244cc86} & 1 & no-think, sampled, think \\
\texttt{Qwen/Qwen3-4B} & \texttt{1cfa9a7208} & 1 & no-think, sampled, think \\
\texttt{Qwen/Qwen3-8B} & \texttt{b968826d9c} & 1 & no-think, sampled, think \\
\texttt{Qwen/Qwen3-14B} & \texttt{40c069824f} & 1 & no-think, sampled, think \\
\texttt{Qwen/Qwen3-32B} & \texttt{9216db5781} & 1 & no-think, sampled, think \\
\texttt{Qwen/Qwen3.6-35B-A3B} & \texttt{995ad96eac} & 1 & no-think, sampled, think \\
\texttt{Qwen/Qwen3.8-27B} & \texttt{1d4bf0f2ff} & 1 & no-think, sampled, think \\
\texttt{Qwen/Qwen3.8-Flash-Next-FP8} & \texttt{236dfdf285} & 4 & no-think, sampled, think \\
\texttt{google/gemma-4-31B-it} & \texttt{842da3794e} & 2 & no-think, sampled \\
\texttt{allenai/Olmo-3.1-32B-Instruct} & \texttt{ac0587e4a7} & 1 & no-think, sampled \\
\bottomrule
\end{tabular}
\caption{Model checkpoints. ``sampled'': ten samples at $T=0.7$ without thinking (consistency analysis). Excluded thinking runs (Gemma-4, OLMo-3.1-Think) are listed in Appendix~\ref{app:deviations}.}
\label{tab:models}
\end{table*}

\paragraph{Decoding and sampling.}
Without thinking, sampling at temperature 0.7 instead of greedy decoding changes closed-book accuracy by $-5.0$ to $+0.3$ points across the eleven models, far less than the thinking effects in the main text. Items are generated with a fixed seed; \cat{element_absence} oversamples charts in which an element is absent on the surface but present among hidden stems. Tables of released results give Wilson intervals \citep{wilson1927probable}. Thinking runs use each model's released sampling configuration (temperature 0.6 for Qwen3, 1.0 for Qwen3.6 and Qwen3.8; top-$p$ 0.95, top-$k$ 20). One sample is drawn per item and condition, except in the consistency analysis, which draws ten samples at temperature 0.7 without thinking; replacing greedy decoding by this sampling changes day-pillar accuracy by $-2.6$ to $+1.3$ points across the eleven models. To estimate sampling variation of the thinking runs, four models draw two further samples (seed 20261101) on the day-pillar items (\cond{closed}, \cond{table}) and the collision \cond{+full} items; across the resulting three samples, cell accuracy varies by a median of 5 points. The largest range, 27 points, is Qwen3.8-Flash-Next's closed-book day pillar (61--88\%), a cell in which many outputs reach the 8k cap. OLMo-3.1-32B-Think reached the cap on 400 of 2,046 requests; among the included thinking runs, every non-truncated output conformed to the schema.

Inference ran on NVIDIA A100 and H100 GPUs with vLLM 0.32.0 in a container. All raw outputs (prompt, model text, parsed answer, grade, token counts) are stored as JSONL. Every table in this paper is generated from the result files by \texttt{scripts/make\_tables.py}, and every number quoted in the text by \texttt{scripts/key\_numbers.py}. Code, data and raw outputs are provided as anonymised supplementary material.

\end{document}